\documentclass[10pt,a4paper,english]{article}
\usepackage{times}

\usepackage{setspace}

\usepackage{amsmath}
\usepackage{amsfonts}
\usepackage{amssymb}
\usepackage{bbm}

\usepackage{algorithm}
\usepackage[noend]{algorithmic}

\usepackage{booktabs}
\usepackage{rotating}
\usepackage{multirow}
\usepackage{array}

\usepackage{graphicx}
\usepackage{subfig}
\usepackage{epstopdf}
\DeclareGraphicsExtensions{.eps,.png,.jpg,.pdf,.tif,.tiff,.ps}
\graphicspath{{./}{img/}{img/raster/}{img/vector/}}

\usepackage{url}
\usepackage{hyperref}
\usepackage[svgnames]{xcolor}

\usepackage{xspace}
\newcommand{\KMeans}{\mbox{K-Means}}%
\newcommand{\FMeasure}{\mbox{$F_{score}$}}%
\hypersetup
{
pdftitle={Semantic-enriched Visual Vocabulary Construction in a Weakly Supervised Context},%
pdfauthor={Marian-Andrei Rizoiu, Julien Velcin and St\'ephane Lallich},%
pdfkeywords={bag-of-features representation;visual vocabulary construction;semantic-enriched representation;semi-supervised learning;feature filtering;representations;heuristic methods;pattern analysis.}%
}

\newcommand{\HRule}{\rule{\linewidth}{0.5mm}}

\begin{document}

\doublespacing


\begin{center}

\vspace*{-4cm}

\HRule \\[-0.5cm]
{ \huge \textbf{Semantic-enriched}\\[-0.8cm]
\textbf{Visual Vocabulary Construction in a}\\%
\textbf{Weakly Supervised Context}}\\

\HRule \\[-1cm]

\begin{center}

\large
Marian-Andrei \textsc{Rizoiu}\footnote{Corresponding author. \\ ERIC Laboratory, University Lumi\`ere Lyon 2  \\ Address: 5, avenue Pierre Mend\`es France, 69676 Bron Cedex, France \\ Tel. +33 (0)4 78 77 31 54 Fax. +33 (0)4 78 77 23 75} \hspace{3cm} Julien \textsc{Velcin} \\
St\'ephane \textsc{Lallich}
 
\end{center}
\vspace*{-0.2cm}

\small
\emph{\{firstname.lastname\}@univ-lyon2.fr} \\

%

\textbf{Abstract}
\end{center}

One of the prevalent learning tasks involving images is content-based image classification.
This is a difficult task especially because the low-level features used to digitally describe images usually capture little information about the semantics of the images.
In this paper, we tackle this difficulty by enriching the semantic content of the image representation by using external knowledge.
The underlying hypothesis of our work is that creating a more semantically rich representation for images would yield higher machine learning performances, without the need to modify the learning algorithms themselves.
The external semantic information is presented under the form of non-positional image labels, therefore positioning our work in a weakly supervised context.
Two approaches are proposed: the first one leverages the labels into the visual vocabulary construction algorithm, the result being dedicated visual vocabularies.
The second approach adds a filtering phase as a pre-processing of the vocabulary construction.
Known positive and known negative sets are constructed and features that are unlikely to be associated with the objects denoted by the labels are filtered.
We apply our proposition to the task of content-based image classification and we show that semantically enriching the image representation yields higher classification performances than the baseline representation.

\begin{center}
\textbf{Keywords}
\end{center}

bag-of-features representation;
visual vocabulary construction; 
semantic-enriched representation;
image numerical representation;
semi-supervised learning.

\newpage
\singlespacing 

\section{Introduction}
\label{sec:introduction}

The large scale production of image data has been facilitated in modern days by the maturing of the image acquisition, storing, transmission and reproduction devices and techniques.
The Web 2.0 allowed easy image sharing and recently even search capabilities (\textit{e.g.}, Instagram\footnote{\url{http://instagram.com/}}, Flickr\footnote{\url{http://www.flickr.com/}}).
Social Networks rely heavily on image sharing.
Because of the sheer volumes of created images, automatic summarization, search and classification methods are required.

The difficulty when analyzing images comes from the fact that digital image numerical formats poorly embed the needed semantic information.
For example, images acquired using a digital photo camera are most often stored in raster format, based on pixels.
A pixel is an atomic image element, which has several characteristics, the most important being the size (as small as possible) and its color.
Other information can be color coding, alpha channel \textit{etc.}.
Therefore, an image is stored numerically as a matrix of pixels.
The difficulty raises from the fact that low-level features, such as position and color of individual pixels, do not capture too much information about the semantic content of the image (\textit{e.g.}, shapes, objects).
This problem is also known as the \textit{semantic gap} between the numerical representation of the image and its intended semantics.
To address this issue, multiple representation paradigms have been proposed, some of which will be presented in Section~\ref{sec:related-work}.
The one showing the most promising results is the ``bag-of-features'' representation, a representation inspired from the textual ``bag-of-words'' textual representation.
Whatsoever, the results obtained by the state-of-the-art image representations still leave plenty of room for improvements.
One of the privileged tracks to closing the semantic gap is to take into account additional information stored in other types of data (\textit{e.g.}, text, labels, ontologies of concepts) associated with the images.
With today's Web, additional information of this type is often available, usually created by anonymous contributors.
Our work presented in this paper is targeted towards improving a baseline, unsupervised, image description strategy by rendering it semi-supervised, in order to take into account user-generated additional information.
The purpose is to capture more of the semantics of an image in its numerical description and to improve the performances of an image-related machine learning task.

\paragraph{An overview of our proposals}
\label{subsec:introduction-overview-proposals}

The focus of the work is embedding semantic information into the construction of image numerical representation.
The task of content-based image classification is used only to assess the validity of our proposals.
The content-based image classification literature provides examples (some of which are mentioned in Section~\ref{sec:related-work}) of systems which achieve good results.
Our objective is not to compare with these approaches or show the superiority of our methods on well-known image benchmarks.
Neither we do not propose a new image representation system.
The objective of our work is to show how embedding semantics into an existing image representation can be beneficial for a learning task, in this case image classification.
Starting from the  baseline image representation construction present in Section~\ref{subsec:introduction-baseline-approach}, we propose two algorithms that make use of external information to enrich the semantics of the image representation.
The external information is under the form of non-positional labels, which signal the presence in the image of an object (\textit{e.g.}, car, motorcycle) or give information about the context of the image (\textit{e.g.}, holiday, evening), but do not give any information about its position of the image (in the case of objects).
Furthermore, the labels are available only for a part of the image collection, therefore positioning our work in a semi-supervised learning context.
We use both the baseline representation and our semantically improved representation in an image classification task and we show that leveraging semantics consistently provides higher scores. 

Our work is focused on the \textit{visual vocabulary} construction (which is also referred in the literature as \textit{codebook} or \textit{model}).
In the \textit{``bag-of-features''} (\emph{BoF}) representation, the visual words serve a similar role as the real textual words do in the ``bag-of-words'' representation.
We propose two novel contributions that leverage external semantic information and that allow the visual vocabulary to capture more accurately the semantics behind a collection of images.
The first proposal deals with introducing the provided additional information early in the creation of the visual vocabulary.
A \textit{dedicated visual vocabulary} is constructed starting from the visual features sampled from images labeled with a given label.
Therefore, a dedicated vocabulary contains visual words adapted to describing the object denoted by the given label.
In the end, the complete visual vocabulary is created by merging the dedicated vocabularies.
In the second proposal, we add a filtering phase as a pre-processing of the visual vocabulary construction.
This reduces the influence of irrelevant features on the visual vocabulary construction, thus enabling the latter to be more adapted to describe the semantics of the collection for images.
For any given image, we construct a known positive set (images labeled with the same labels as the given image) and a known negative set (images that do not share any labels with the given image).
If a visual feature, sampled from the target image, is more similar to features in the known negative set than to features in the known positive set, then there are high chances that it does not belong to the objects denoted by the labels of the given image and it can, therefore, be eliminated.
As our experiments in Section~\ref{subsec:experiments-quantitative-evaluation} show, this approach increases the overall accuracy of the image-related learning task.
The two approaches are combined into a visual vocabulary construction technique and shown to consistently provide better performances than the baseline technique presented in Section~\ref{subsec:introduction-baseline-approach}.

\paragraph{The layout of this article}
The remainder of this paper is structured as follows: the rest of this section presents how to construct a baseline ``bag-of-features'' image description (in Section~\ref{subsec:introduction-baseline-approach}).
In Section~\ref{sec:related-work}, we present a brief overview on constructing a numerical image representation, concentrating on some of the state-of-the-art papers that relate to visual vocabulary construction and knowledge injection into image representation. 
Section~\ref{sec:proposals} explains our two proposals, followed, in Section~\ref{sec:experiments}, by the experiments that were performed. 
Some conclusions are drawn and future work perspectives are given in Section~\ref{sec:conclusions}.

\subsection{Baseline ``bag-of-features'' image numerical description}
\label{subsec:introduction-baseline-approach}

The ``bag-of-features''~\cite{CSU04,ZHA07a} (\emph{BoF}) representation is an image representation inspired from the ``bag-of-words'' (\emph{BoW}) textual representation.
The \emph{BoW} representation is an orderless document representation, in which each document is depicted by a vector of frequencies of words over a given dictionary.
\emph{BoF} models have proven to be effective for object classification~\cite{CSU04,WIL04}, unsupervised discovery of categories~\cite{FEI05,QUA05,SIV05} and video retrieval~\cite{CAM08,SIV03}.
For object recognition tasks, local features play the role of ``visual words'', being predictive of a certain ``topic'' or object class.
For example, a wheal is highly predictive of a bike being present in the image.
If the visual dictionary contains words that are sufficiently discriminative when taken individually, then it is possible to achieve a high degree of success for whole image classification.
The identification of the object class contained in the image is possible without attempting to segment or localize that object, simply by looking which visual words are present, regardless of their spatial layout.
Overall, there is an emerging consensus in recent literature that \emph{BoF} methods are effective for image description~\cite{ZHA07a}.

\begin{figure}[htb]
	\centering
	\includegraphics[width=0.99\textwidth]{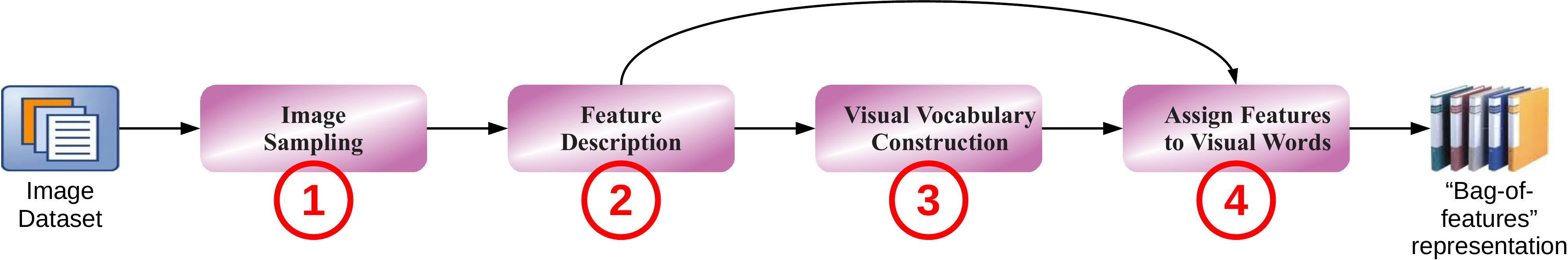}
	
	\caption{Construction flow of a ``bag-of-features'' numerical representation for images}
	\label{fig:bag-of-features}
\end{figure}

\paragraph{Baseline construction}
Typically, constructing a \emph{BoF} image representation is a four phase process, as shown in Figure~\ref{fig:bag-of-features}.
Starting from a collection $\mathcal{P}$ containing $n$ images, the purpose is to translate the images into a numerical space, in which the learning algorithm is efficient.
In \textit{phase~1}, each image $p_i \in \mathcal{P}$ is sampled and $l_i$ patches (features)\footnote{$l_i$ is dependent on the content on the image (number of objects, shape \textit{etc.}) and the extraction algorithm used. It can vary from a couple of hundreds of features up to several tens of thousands. } are extracted.
Many sampling techniques have been proposed, the most popular being dense grid sampling~\cite{FEI05,VOG07} and salient keypoint detector~\cite{CSU04,FEI05,SIV05}.
In \textit{phase~2}, using a local descriptor, each feature is described using a $h$-dimensional\footnote{\textit{e.g.} for the SIFT descriptor $h = 128$.} vector.
The SIFT~\cite{LOW04} and the SURF~\cite{BAY06} descriptors are popular choices.
Therefore, after this phase, each image $ p_i $ is numerically described by $V_i \subset \mathbb{R}^{h}$ , the set of $h$-dimensional vectors describing features sampled from $p_i$.

\begin{figure}[!t]
  \centering
  \subfloat[] {
  	\label{subfig:example-bike-1}
  	\includegraphics[height=0.124\textheight]{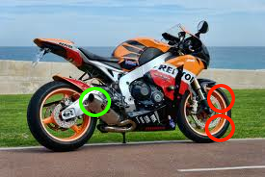}
  }         
  \hfill
  \subfloat[]{
  	\label{subfig:example-bike-2}
  	\includegraphics[height=0.124\textheight]{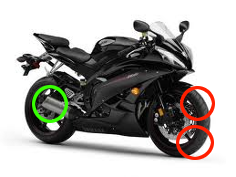}
  }
  \hfill
  \subfloat[]{
  	\label{subfig:example-bike-3}
  	\includegraphics[height=0.124\textheight]{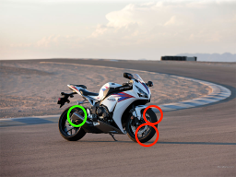}
  }
  
  \caption{Example of feature corresponding to the visual words associated with ``wheal'' (in red) and ``exhaust pipe'' (in green)}
  \label{fig:visual-words-example}
\end{figure}

Based on these numeric features, in \textit{phase~3}, a visual vocabulary is constructed using, for example, one of the techniques presented in Section~\ref{subsec:related-works-visual-vocabulary-techniques}.
This is usually achieved by means of clustering of the described features, and the choice is usually the \KMeans{} clustering algorithm, for its linear execution time required by the high number of features.
The visual vocabulary is a collection of $m$ visual words, which are described in the same numerical space as the features and which serve as the bases of the numerical space in which the images are translated.
More precisely, the centroids created by the clustering algorithm serve as visual words.
In clustering, centroids are the abstractions of a group of documents, therefore summarizing the common part of the documents.
In the above example, all the visual features extracted from the region of an image depicting the wheal of a bike  will be regrouped together into one or several clusters.
The centroid of each cluster represents a visual word, which is associated with the wheal.
Figure~\ref{fig:visual-words-example}, we depict three examples of images portraying bikes.
In each image, we highlight 3 features: two corresponding to visual words associated with ``wheal'' and one associated with a visual word associated with ''exhaust pipe''.

In \textit{phase~4}, each sampled feature is assigned to a visual word.
Similarly to the \emph{BoW} numerical description for texts, each image is described as a distribution over the visual words, using one of the term weighting scheme (\textit{e.g.}, $tf$, $tf \text{x} idf$ \textit{etc.}).
In the previous example, the distribution vector associated with each of the images in Figure~\ref{fig:visual-words-example} has a high count for the visual words associated with ``wheal'', ``exhaust pipe'', and ``sadle''.
The resulting numerical description can then be used for classification, information retrieval or indexation tasks.

\section{Context and related work}
\label{sec:related-work}

Over the past decades computer vision domain has seen a large interest from the research community.
Its application are larger than image analysis and include augmented reality, robotic vision, gesture recognition \textit{etc.}
Whatsoever, in the context of Internet-originating images, one of the prevailing task is content-based image classification.
Some of the initial image classification systems used color histograms~\cite{SWA91} for image representation.
Such a representation does not retain any information about the shapes of objects in images and obtains moderate results.
Other systems~\cite{HAR73,LAZ03,MED02,VAR03} rely on texture detection.
Texture is characterized by the repetition of basic elements or \emph{textons}.
For stochastic textures, it is the identity of the textons, not their spatial arrangement, that matters.
The \emph{BoF} orderless representation has imposed itself as the state-of-the-art in image representation, for classification and indexation purposes.
The process of constructing the representation includes sampling the image (\textit{phase~1} in Figure~\ref{fig:bag-of-features}), describing each features using an appearance-based descriptor (\textit{phase~2}), constructing a visual vocabulary (\textit{phase~3}) and describing images as histograms over the visual words (\textit{phase~4}).

The remainder of this section presents a brief overview (i) of the sampling strategies and numerical descriptors for image keypoints present in literature (in Section~\ref{subsec:related-works-overview-bag-of-features}) and (ii) of the visual vocabulary construction techniques, concentrating on how external information can be used to improve the vocabularies representativity (in Section~\ref{subsec:related-works-visual-vocabulary-techniques}).

\subsection{Sampling strategies and numerical description of image features}
\label{subsec:related-works-overview-bag-of-features}

\paragraph{Image sampling methods}

Image sampling for the \emph{BoF} representation is the process of deciding which regions of a given image should be numerically described.
In Figure~\ref{fig:bag-of-features}, it corresponds to \textit{phase~1} of the construction of a \emph{BoF} numerical representation.
The output of feature detection is a set of patches, identified by their locations in the image and their corresponding scales and orientations.
Multiple sampling methods exist~\cite{OHA11}, including \textit{Interest Point Operators}, \textit{Visual Saliency} and random or dense grid sampling.

\textit{Interest Point Operators}~\cite{KAD01,LOW99} search to find patches that are stable under minor affine and photometric transformations.
Interest point operators detect locally discriminating features, such as corners, blob-like regions, or curves. 
A filter is used to detect these features, measuring the responses in a three dimensional space.
Extreme values for the responses are considered as interest points.
The popular choice is the Harris-Affine detector~\cite{MIK04}, which uses a scale space representation with oriented elliptical regions.
\textit{Visual Saliency}~\cite{FRIN10} feature detectors are based on \textit{biomimetic} computational models of the human visual attention system.
Less used by the \emph{BoF} literature, these methods are concerned with finding locations in images that are visually salient. 
In this case, fitness is often measured by how well the computational methods predict human eye fixations recorded by an eye tracker. 
There are research~\cite{SIV03} that argue that interest point-based patch sampling, while useful for image alignment, is not adapted for image classification tasks.
Examples are city images, for which the interest point detector does not consider relevant most of the concrete and asphalt surroundings, but which are good indicators of the images' semantics.
Some approaches sample patches by using \textit{random sampling}~\cite{MAR05}.
\cite{NOW06} compare a random sampler with two interest point detectors: \textit{Laplacian of Gaussian}~\cite{LIN93} and \textit{Harris-Laplace}~\cite{LAZ03a}.
They show that, when using enough samples, random sampling exceeds the performance of interest point operators.
\textit{Spatial Pyramid Matching}, proposed in \cite{LAZ06}, introduces spacial information in the orderless \emph{BoF} representation by creating a pyramid representation, where each level divides the image in increasingly small regions.
Feature histogram is calculated for each of these regions.
The distance between two images using this spatial pyramid representation is a weighted histogram intersection function, where weights are largest for the smallest regions.

\paragraph{Feature descriptors}

With the image sampled and a set of patches extracted, the next questions is how to numerically represent the neighborhood of pixels near a localized region.
In Figure~\ref{fig:bag-of-features}, this corresponds to \textit{phase~2} of the construction of a \emph{BoF} numerical representation.
Initial feature descriptors simply used the pixel intensity values, scaled for the size of the region.
The \textit{normalized pixel values} have been shown~\cite{FEI05} to be outperformed by more sophisticated feature descriptors, such as the SIFT descriptor.
The \textit{SIFT} (Scale Invariant Feature Transform)~\cite{LOW04} descriptor is today's most widely used descriptor.
The responses to 8 gradient orientations at each of 16 cells of a 4x4 grid generate the 128 components of the description vector.
Alternative have been proposed, such as the \textit{SURF} (Speeded Up Robust Features)~\cite{BAY06} descriptor. 
The SURF algorithm contains both feature detection and description.
It is designed to speed up the process of creating features similar to those produced by a SIFT descriptor on Hessian-Laplace interest points by using efficient approximations.

\subsection{Unsupervised visual vocabulary construction}
\label{subsec:related-works-visual-vocabulary-techniques}

The visual vocabulary is a mid-level transition key between the low-level features and a high-level representation.
It is a prototypic representation of features that are discriminative in a classification context.

The visual vocabulary is used to reduce dimensionality and to create a fixed length numerical representation for all images\footnote{The number of extracted features can greatly vary depending on the image and the method used for sampling.}.
Most \emph{BoF} approaches use clustering to created the visual vocabulary, usually the \KMeans{}~\cite{JIA07,LAZ06,SIV03} algorithm.
\KMeans{} is used for the fact that it produces centroids, which are prototypes of similar features in the same cluster. 
Its linear execution time is a plus considering the high volume of individuals to be processed~\cite{PHA08}.
Some authors~\cite{JUR05} argument that in \KMeans{}, centroids are attracted by dense regions and under-represent less denser, but equally informative regions.
Therefore, methods were proposed for allocating centers more uniformly, inspired by mean shift~\cite{COM02} and on-line facility location~\cite{MEY01}.
Other visual vocabulary construction techniques do not rely on \KMeans{}.
For example, \cite{MOO07} use an \textit{Extremely Randomized Clustering Forest}, an ensemble of randomly created clustering trees.
This technique provides good resistance to background clutter, but the main advantage over \KMeans{} is the faster training time.

One of the most important parameters in the construction of the visual vocabulary is its dimension, which has a powerful impact on both performance and computational complexity~\cite{CSU04,JUR05}.
It has been shown~\cite{JIA07,LOP10,NOW06} that a large vocabulary may lead to overfitting for construction techniques based on interest points detection.
As our experiments show (in Section~\ref{subsec:experiments-overfitting}), even a random vocabulary (in a random vocabulary, a number of features are randomly chosen to serve as visual words) can lead to overfitting if its dimension is high enough.

\subsection{Leveraging additional information}
\label{subsec:related-works-additional-information}

The \emph{BoF} representation yields surprising results for image classification and indexing.
This is because there is an intrinsic relation between the ``quantity'' of semantic information captured by the description space and the performances of machine learning algorithms (\textit{e.g.}, in a classification task, the separability of individuals in the description space is crucial).
Therefore, one direction to further improve results is to construct new representations that capture even more semantics from the raw image data.
Another direction, the one that we privilege in our work, is to use external information to further enrich the semantic content of the constructed representation.
In the case of Internet-originating images, precious information is given either by the textual context of images (\textit{e.g.}, titles, descriptions \textit{etc.}), or by labels attached to the images (\textit{e.g.}, on social networks websites, users have the option to label the presence of their friends in images).
Of course, the literature presents approaches that leverage other resources to semantically enrich enrich the image representation (\textit{e.g.}, \cite{ATH05} propose a system that links low-level visual descriptors to high-level, domain-specific concepts in an ontology).
In the following paragraphs, we detail some of the methods present in the literature that address the use of additional information under the form of text or labels in order to improve image classification results and we position our work relative to these approaches.

\paragraph{Leveraging the image's textual context}
In~\cite{MOR09}, the text that comes alongside the images is used to improve the visual query accuracy.
A \emph{BoF} representation for images is created as shown in Section~\ref{subsec:introduction-baseline-approach}, with the exception that color information is also added to the keypoint description.
An 11-dimension vector coding the color information of the sampled patches is added to the 128-dimension vector generated by the SIFT.
The text that surrounds the images in the web pages is used to extract topics, using LDA~\cite{BLE03}.
The inferred topics are, afterwards, used to describe the textual information (therefore functioning as a dimension reduction technique).
The textual and the image data are used together to estimate the parameters of a probabilistic graphical model, which is trained using a small quantity of labeled data.
Another approach that uses the text accompanying images originating from the Internet is presented in~\cite{WAN09}.
An auxiliary collection of Internet-originating images, with text attached, is used to create a textual description of a target image.
Images are described using three types of features: the SIFT features, the GIST features~\cite{OLI01} and local patch color information.
For each test image, the $K$ most similar images (in terms of visual features) are identified in the auxiliary collection.
The text associated with these near neighbor images is summarized to build the text feature.
The label of each image is considered as a unit (\textit{i.e.}, a whole phrase is considered as an item) and the text feature is constructed as a normalized histogram over labels.
A text classifier and a visual classifier are trained and the outputs of the two classifiers are merged for a more accurate description of the photo.
\cite{MOO08} use co-training~\cite{BLU98} to construct a classifier starting from textual and visual data.
Text is described using a \emph{BoW} representation, whereas images are described using region-based features.
Each image is divided into a number of regions of fixed dimension (4-by-6 pixels), which are described using texture and color features.
Co-training is a semi-supervised classification technique, which first learns a separate classifier for textual data and image data, using any labeled examples. 
The most confident predictions of each classifier on the unlabeled data are then used to iteratively construct additional labeled training data, and the classifiers are re-trained.

\paragraph{Leveraging external semantic knowledge}
Other solutions rely on external expert knowledge in order to guide the visual vocabulary construction.
This knowledge is most often expressed under the form of class/category annotations or labels (\textit{e.g.} signaling the presence of an object inside an image), or semantic resources, such as WordNet~\cite{MIL95}.
An iterative boosting-like approach is used in \cite{ZHA09}.
Each iteration of boosting begins by
learning a visual vocabulary according to the weights assigned by the previous boosting iteration. 
The resulting visual vocabulary is then applied to encode the training examples, a new classifier is learned and new weights are computed.
The visual vocabulary is learned by clustering using \KMeans{} a ``learning'' subset of image features.
Features from images with high weights have more chances of being part of the learning subset.
To classify a new example, the AdaBoost~\cite{FRE97} weighted voting scheme is used.

\cite{PER06} construct both a generic vocabulary and a specific one for each class.
The generic vocabulary describes the content of all the considered classes of images, while the specific vocabularies are obtained through the adaptation of the universal vocabulary using class-specific data. 
Any given image can, afterwards, be described using the generic vocabulary or one of the class-specific vocabularies.
A semi-supervised technique~\cite{JI10}, based on Hidden Random Markov Fields, uses local features as Observed Fields and Semantic labels as Hidden Fields and employs WordNet to make correlations.
Some works~\cite{FUL08,HSU05,LAZ09,WIN05} use mutual information between features and class labels in order to learn class-specific vocabularies, by merging or splitting initial visual words quantized by \KMeans{}.
Another work~\cite{LIU09} presents an algorithm used for learning a generic visual vocabulary, while trying to preserve and use the semantic information in the form of a point-wise mutual information vector. 
It uses the diffusion distance to measure intrinsic geometric relations between features.
Other approaches~\cite{MAR06} make use of label positioning in the images to distinguish between foreground and background features. 
They use weights for features, higher for the ones corresponding to objects and lower for the background.

\paragraph{Our positioning}
In the methods presented earlier, we identify several approaches towards improving the results of classification algorithms: (a) improving image representation semantics by combining multiple types of visual features (\textit{e.g.}, SIFT, color, texture \textit{etc.}, no external information is leveraged), (b) modifying the classification algorithm to take into account the text/label information (usually by training separate classifiers for (i) text and image or (ii) based on each label), (c) training and using multiple vocabularies to describe an image and (d) making use of positional labels to filter features unlikely to be relevant. 
Positional labels are labels in which the position of the objects in images are known, in addition to their presence.
This kind of labeling is usually more costly to perform than non-positional labeling.

Our proposals deal with leveraging external information to enrich the semantics of the image representation.
The additional information is taken into account at the level of the representation construction. 
We do not modify the learning algorithm, therefore our proposals are compatible with existing classification algorithm.
Our proposals can be classified under the previously defined point (c), since we construct multiple \emph{dedicated visual vocabularies}.
To the best or our knowledge, the feature filtering proposal, detailed in Section~\ref{subsec:proposals-filtering-points-interest}, is the first algorithm aimed at filtering features irrelevant for an object, without making use of positional labels.
This greatly reduces the effort of manually labeling and allows the usage of broader sources of data available on the Web.

\section{Improving the \emph{BoF} representation using semantic knowledge}
\label{sec:proposals}

In this section, we present our two novel methods that leverage external semantic information, under the form of non-positional \textit{object labels}, into the visual vocabulary construction.
This kind of information is often freely available on the Web, being constantly produced by anonymous users.
There are some distinctive differences between using public domain knowledge and using domain experts:
(i) the public domain knowledge is not created especially for our application, but rather we make use it to improve the quality of the constructed image representations and
(ii) unlike using domain experts, no evaluation feedback link can be made back to the creators of additional knowledge, as they are often anonymous.
For these reasons, we privilege a semi-supervised approach and we use the additional information to guide the algorithm in the solutions space.

Our work is positioned in a weakly supervised context, similar to the one defined in~\cite{ZHA07a}.
Each label signals the presence of a given object in an image, but not its position or boundaries.
Our approaches use the semantic information to increase the relevancy of the visual vocabulary.
In our first approach, for each label, we construct a \textit{dedicated visual vocabulary}, based only on the images with a certain label.
Such approaches have been shown~\cite{JIA11,PER06} to improve accuracy over a general purpose vocabulary, since specialized vocabularies contain visual words that more appropriately describe the objects appearing in the image collection.
In our second approach, we further improve accuracy by proposing a novel pre-processing phase, which filters out features that are unlikely to belong to the respective object.
Our filtering proposal follows the framework of the object recognition algorithm proposed in~\cite{LOW04} and uses a positive and a negative example set, constructed based on the labels.
The filtering pre-processing is combined with the \textit{dedicated visual vocabulary} construction, and we show in Section~\ref{sec:experiments} that this approach consistently achieves higher accuracy then both a dedicated vocabulary (with no filtering) and a general purpose vocabulary.

\paragraph{Including semantic knowledge}
The semantic knowledge is presented under the form of a collection $\mathcal{T}$ of $k$ labels, $\mathcal{T} = \{t_{i} | i = 1,2...k\}$.
Each label is considered to denote an object in the image (\textit{e.g.}, a car, a person, a tree), but no positional markers are available.
We make the assumption that the objects denoted by labels do not overlap in the images and their appearance in the dataset is not correlated (\textit{e.g.}, if a car appears, it does not necessarily mean that there is a person next to it).
While these are strong assumptions, we will discuss ways of relaxing them in Section~\ref{sec:conclusions}.
Furthermore, we consider the labeling to be complete (\textit{i.e.}, if an image does not have a given label, than the object does not appear in the image).
In Section~\ref{subsec:proposals-filtering-points-interest}, we discuss in further detail the effects of incomplete labeling, after presenting our proposals.

Only a fraction of the image dataset is labeled and we use both labeled and unlabeled images to construct the semantic-aware representation, therefore positioning our work in the domain of semi-supervised learning.
We denote by $\mathcal{P}$ the input collection, having $n$ images.
$n_1$ images are labeled, thus forming the labeled set ($\mathcal{P}_1$), while the remaining images have no labels.
The \textit{a priori} label information is presented in the form of a boolean matrix $Y \in \{0,1\} ^{n_1 \times k}$, having $n_1$ lines and $k$ columns so that
\begin{equation*}
	y_{i,j} = 
	\begin{cases}
		1 & \text{if image } p_i \in \mathcal{P}_1 \text{ is labeled using } t_j \text{;}\\
    0 & \text{otherwise.}
	\end{cases}
\end{equation*}

\subsection{Dedicated visual vocabulary generation}
\label{subsec:proposals-label-based-model-generation}

The idea behind the \emph{BoF} representation is that the visual words are predictive for certain objects (as seen in Section~\ref{subsec:introduction-baseline-approach}).
The quality of the visual words (and their predictive power) would be enhanced if they are constructed starting only from the features extracted from the respective objects.
This would eliminate the background originating features and features belonging to other objects.
In a weakly supervised context, the object boundaries are unknown, but selecting only the images that contain a certain object increases the relevant/noise feature ratio.
Consequently, the resulted visual words are more accurate descriptions of the objects denoted by the labels.
We propose to construct a \textit{dedicated visual vocabulary} for each label $t_i \in \mathcal{T}$, starting only from features extracted from the images labeled with $t_i$.

\begin{algorithm}[t]                      
\caption{Dedicated vocabulary generation algorithm.}     
\label{algo:dedicated-visual-vocabulary-creationg}  
\begin{algorithmic}                 
\REQUIRE $ \mathcal{C} = \{V_i \: | \: i=1,2..n_1 \}$ - set of features sampled from labeled images
\REQUIRE $Y \in \{0,1\} ^{n_1 \times k}$ - image/label association matrix
\REQUIRE $m$ - the dimension of the visual vocabulary $M$
\ENSURE the visual vocabulary $M$ having $m$ visual words

\STATE \textit{// for each label}
\FOR{ $ i = 1 $ to $k$}	
	\STATE $m_{i} \gets m/k $  \hspace{10mm} \textit{// size of the dedicated vocabulary}
	
	\STATE $ \mathcal{C}_i = \bigcup^{n_1}_{j = 1} V_j \, | \, y_{j,i} = 1 $ \hspace{5mm} \textit{// set of features in images labeled with $t_i$}
	\STATE \textit{// construct dedicated visual vocabulary $M_i$}
	\STATE $M_i \gets $ \textbf{choose\_features\_at\_random} ($m_{i}$, $\mathcal{C}_i$)
	\STATE $M_i \gets $ \textbf{ameliorate\_using\_\KMeans{}} ($M_i,\, \mathcal{C}_i$)
\ENDFOR

\STATE \textit{// merge the dedicated visual vocabularies}
\STATE $ M \gets \emptyset$
\FOR{ $ i = 1 $ to $k$}
	\STATE $M \gets $ \textbf{concatenate\_vocabularies}($M$, $M_i$)
\ENDFOR

\end{algorithmic}
\end{algorithm}

The proposed method is presented in Algorithm~\ref{algo:dedicated-visual-vocabulary-creationg}.
We make no assumptions about the number of visual words needed to describe each object and, therefore, visual words are distributed equally among objects.
We construct $k$ dedicated vocabularies, each one containing $ m/k$ visual words.
Other division techniques can be imagined and make part of the perspectives of our work.
Each dedicated vocabulary is created in the standard \emph{BoF} approach, shown in Section~\ref{subsec:introduction-baseline-approach}.
For a given label $t_i$, we create $\mathcal{C}_i$, the collection of all the features extracted from images labeled with $t_i$.
Formally:
\begin{equation*}
	\mathcal{C}_i = \bigcup^{n_1}_{\substack{j = 1 \\ y_{j,i} = 1}} V_j
\end{equation*}
where $V_j$ is the set of numerically described features sampled from image $p_j$.
The function \textbf{choose\_features\_at\_random} is used to initialize the dedicated vocabulary $M_i$ with $m/k$ features randomly picked from $\mathcal{C}_i$.
The function \textbf{ameliorate\_using\_\KMeans{}} evolves the visual vocabulary $M_i$ by clustering the features in $\mathcal{C}_i$ around the visual words, using the \KMeans{} algorithm.
The Euclidean distance is used to measure the similarity between the numeric descriptions of two features, since this is the distance employed in the original work~\cite{LOW04} that proposed the SIFT descriptor.
Subsequently, it has been used by most of the literature to measure the similarity between features.
The set of resulted visual words represent more accurately the object denoted by the label $t_i$.
At the end of the algorithm, the \textbf{concatenate\_vocabularies} function merges the \textit{dedicated vocabularies} $M_i, i=1,2..k$ into the general visual vocabulary $M$.
This ensures that the generated visual vocabulary contains visual words which describe all the objects labeled with labels in $\mathcal{T}$.

\paragraph{Temporal complexity} Algorithm~\ref{algo:dedicated-visual-vocabulary-creationg} has a linear execution time, if we consider that matrix operations are indivisible and executed in $ O (1) $, which is the case in modern vectorial mathematical environments.
Since we are executing \KMeans{} $k$ times, the temporal complexity will be $ no_{iter} \times k \times O(m/k \times n_{t_i})$, where $n_{t_i}$ is the number of images labeled with $t_i$ and $ no_{iter}$ is the number of performed iterations (usually limited, thus ignored in practice). 
That leads to a theoretical complexity of $ O ( m \times n )$, equal to that of \KMeans{}.

\subsection{Filtering irrelevant features}
\label{subsec:proposals-filtering-points-interest}

We propose a filtering mechanism in order to further increase the relevant/noise features ratio in the dedicated vocabulary construction technique presented in the previous Section~\ref{subsec:proposals-label-based-model-generation}: we detect and filter the features that are unlikely to be related to the object denoted by a given label.
Given an image $p_i \in \mathcal{P}_1$, we construct two auxiliary image collections:
the \textit{known positive set}, which contains only images that are labeled identically as $p_i$, and the \textit{known negative set}, which contains images that do not share any tags with  $p_i$ (given the complete labeling assumption).
In practice, we limit the sizes of the \textit{known positive set} and the \textit{known negative set} to a maximum number of images, given by a parameter $maxFiles$.
We define $KP_{p_i}$ as the set of features sampled from images in the positive set and $KN_{p_i}$ as the set of features sampled from the negative set:
\begin{align*}
	KP_{p_i} &= \{ f^{+} \in V_j \, | \, \forall \: t_l \in \mathcal{T} \text{ for which } y_{i,l} = 1 \Longrightarrow \:y_{j,l} = 1   \} \\
	KN_{p_i} &= \{ f^{-} \in V_j \, | \, \forall \: t_l \in \mathcal{T} \text{ for which } y_{i,l} = 1 \Longrightarrow \:y_{j,l} = 0   \}
\end{align*}
Consider a feature sampled from $p_i$ ($f \in V_i$), which is more similar to the features in the negative collection ($f^{-} \in KN_{p_i}$) rather than the ones in the positive collection ($ f^{+} \in KP_{p_i}$).
Such a feature has a higher chance of belonging to the background of $p_i$ rather than to the objects in the image.
It can, therefore, be filtered.
To measure the similarity of two features, the \textit{euclidean distance} is usually used: $||f_1 - f_2|| = \sqrt{ \Sigma_{i=1}^{h}\left(f_{1,i} - f_{2,i} \right)^{2}}$.
Formally, for a feature $f$ sampled from an image $p_i$:
\begin{align}
	f \in V_i \text{ is filtered } &\Leftrightarrow  \nexists f^{+} \in KP_{p_i} \text{ so that } ||f - f^{+}|| \leq \delta \notag \\
	\text{with } \delta &= \alpha \times \min_{ f \in KN_{p_i}} ||f - f^{-}|| \label{eq:threshold-alpha}
\end{align}
where $\delta$ is the filtering threshold and $\alpha \in \mathbb{R}^{+}$ is a parameter, which allows the fine tuning of the filtering threshold.
The filtering threshold $\delta$ is defined as the distance from the feature $f$ to the closest feature in the known negative set, scaled by tuning parameter $\alpha$.
The influence of parameter $\alpha$ on the effectiveness of the filtering is studied in Section~\ref{subsec:experiments-parameter-alpha}.
A feature $f$ is considered similar to a feature $f^+ \in KP_{p_i}$ if and only if $||f - f^+||$ is lower than the filtering threshold.
Therefore, the feature $f$ is removed when it has no similar feature in the known positive set.

\begin{figure}[!t]
  \centering
  \subfloat[] {
  	\label{subfig:target-image}
  	\includegraphics[height=0.124\textheight]{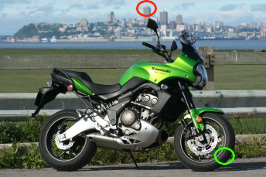}
  }         
  \hfill
  \subfloat[]{
  	\label{subfig:known-positive-set}
  	\includegraphics[height=0.124\textheight]{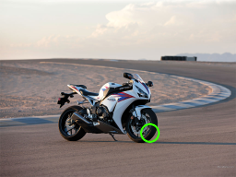}
  }
  \hfill
  \subfloat[]{
  	\label{subfig:known-negative-set}
  	\includegraphics[height=0.124\textheight]{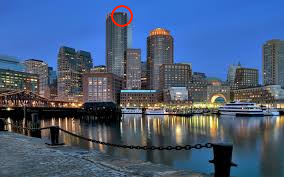}
  }
  
  \caption{ (a) An image labeled ``motorbike'', (b) an image from the \textit{known positive set} and (c) an image from the \textit{known negative set}}
  \label{fig:filtering-example}
\end{figure}

Let's take the example of image collection depicted in Figure~\ref{fig:filtering-example}.
The images in Figures~\ref{subfig:target-image} and~\ref{subfig:known-positive-set} are labeled ``motorbike'', whereas the image in Figure~\ref{subfig:known-negative-set} is labeled ``city''.
The target image in Figure~\ref{subfig:target-image} has buildings in the background, and any feature sampled from that region of the image would be irrelevant for the object motorbike.
Figure~\ref{subfig:known-positive-set} serves as \textit{known positive set}, while Figure~\ref{subfig:known-negative-set} serves as \textit{known negative set}.
We take the example of two features: $f_1$ sampled from the wheal of the motorbike (shown in green) and $f_2$ sampled from the buildings in the background (shown in red), of the target image.
For $f_1$, at least one similar feature exists in the positive set.
For $f_2$, no similar features exist in the known positive set, its most similar feature being in the \textit{known negative set} (shown in red in Figure~\ref{subfig:known-negative-set}).
$f_2$ is, therefore, eliminated as it is considered not relevant for the object motorbike.

Algorithm \ref{algo:point-of-interest-filtering-algorithm} presents the proposed filtering algorithm.
The algorithm has two parameters $maxFiles$, which controls the maximum size of the $KP_{p_i}$ and $KN_{p_i}$ sets, and $\alpha$, which controls how strict is the filtering.
For each labeled image $p_i$, the functions \textbf{create\_KP} and \textbf{create\_KN} are used to construct the feature sets $KP_{p_i}$ and, respectively, $KN_{p_i}$.
The \textbf{count\_similar} function is used to count how many features in $KP_{p_i}$ have the similarity distance lower than the filtering threshold.
If there exists at least one such feature in the $KP_{p_i}$ set, then $f$ is added to $V_i^f$, the filtered feature set of $p_i$.

\begin{algorithm}[t]
	\caption{Filtering irrelevant features.}
	\label{algo:point-of-interest-filtering-algorithm}
	
\begin{algorithmic}                  
\REQUIRE $ \mathcal{C} = \{V_i \: | \: i=1,2..n_1 \}$ - set of features sampled from labeled images
\REQUIRE $ Y \in \{0,1\} ^{n_1 \times k}$ - image/label association matrix
\STATE \textbf{Parameter:} $\alpha$ - parameter controlling the filtering threshold
\STATE \textbf{Parameter:} $maxFiles$ - controls the size of the known positive and known negative sets
\ENSURE $ V_i^f, \: i=1,2..n_1 $ - sets of the filtered features in each labeled image

\STATE \textit{// for each labeled image}
\FOR{ $ i = 1 $ to $n_1$}
	\STATE $ V_i^f \gets \emptyset $
	\STATE $ \mathcal{T}_i \gets \{ t_j \: | \: y_{i,j} = 1 \}$   \hspace{10mm} \textit{// the labels of image $p_i$}

	\STATE $KP_{p_i} \gets $ \textbf{create\_KP}($i$, $T_i$, $Y$, $\mathcal{C}$, $maxFiles$) \hspace{5mm} \textit{// $KnownPositive$ set}
	\STATE $KN_{p_i} \gets $ \textbf{create\_KN}($i$, $T_i$, $Y$, $\mathcal{C}$, $maxFiles$) \hspace{5mm} \textit{// $KnownNegative$ set}

	\STATE \textit{// process each feature in current image $p_i$}
	\FOR{ each $f \in V_i$ }
		\STATE $ \delta \gets \alpha \times$ \textbf{min\_distance}($f$, $KN_{p_i}$)
		\STATE $ count \gets $ \textbf{count\_similar}($f$, $KP_{p_i}$, $\delta$)
		\IF{ $count > 0$ }
			\STATE $V_i^f \gets V_i^f \bigcup \{ f \} $
		\ENDIF
	\ENDFOR
	
\ENDFOR
\end{algorithmic}
\end{algorithm}

\paragraph{Temporal complexity}
In Algorithm~\ref{algo:point-of-interest-filtering-algorithm}, for comprehension reasons, operations are presented for each feature $f$ sampled from the image $p_i$.
In reality, in vectorial mathematical environments (\textit{e.g.} $Octave$), matrix operations are unitary and they can be considered to be executed in $O(1)$.
Thus, the algorithm has a linear execution time $O(n_1 \times maxFiles)$.

\paragraph{Incomplete labeling}
In the proposed approaches, as well as in the experiments presented in Section~\ref{sec:experiments}, we make the assumption of complete labeling: if an object occurs in an image, then it is sure that the image has its corresponding label attached.
In the case of incomplete labeling, an object might appear in an image $p$, but the associate label $t$ is not set for the image $p$.
For the dedicated vocabulary construction, incomplete labeling has a limited impact, especially if the dataset is large enough.
It only means that the image $p$ is left out when constructing the vocabulary for label $t$.
For the filtering proposal, missing labels mean that the image $p$ has a chance of being selected for the \textit{known negative set} for an image labeled with $t$.
This translates into a very high filtering threshold.
Still, this should not pose problems if the \textit{known positive set} also contains images depicting the given object.
A given feature needs to have only one similar feature in the known positive set to be considered representative for the object.
Furthermore, considering that our algorithms are devised to work in a semi-supervised context, a limited number of completely labeled images is required.
This reduces considerably the manual labeling effort.

\section{Experiments and results}
\label{sec:experiments}

As already pointed out in Section~\ref{sec:introduction}, the focus of our work is enriching the semantics of the numerical representation of images.
Therefore, the purpose of the experiments presented in this section is to compare the semantically-enriched representations created by our proposals to a standard baseline representation, created as described in Section~\ref{subsec:introduction-baseline-approach}.
Whatsoever, directly comparing the discriminative power of two representations is not possible, unless in the context of an image-related machine learning task in this case a content-based image classification.
In a nutshell, starting from a collection of images, we construct multiple numerical representations (corresponding to the techniques to be compared) and we train identical classifiers based on each of these representations.
In the end, we attribute the differences of classifier performance as a direct consequence of the representation construction technique.

\begin{figure}[tb]
	\centering
	\includegraphics[width=0.99\textwidth]{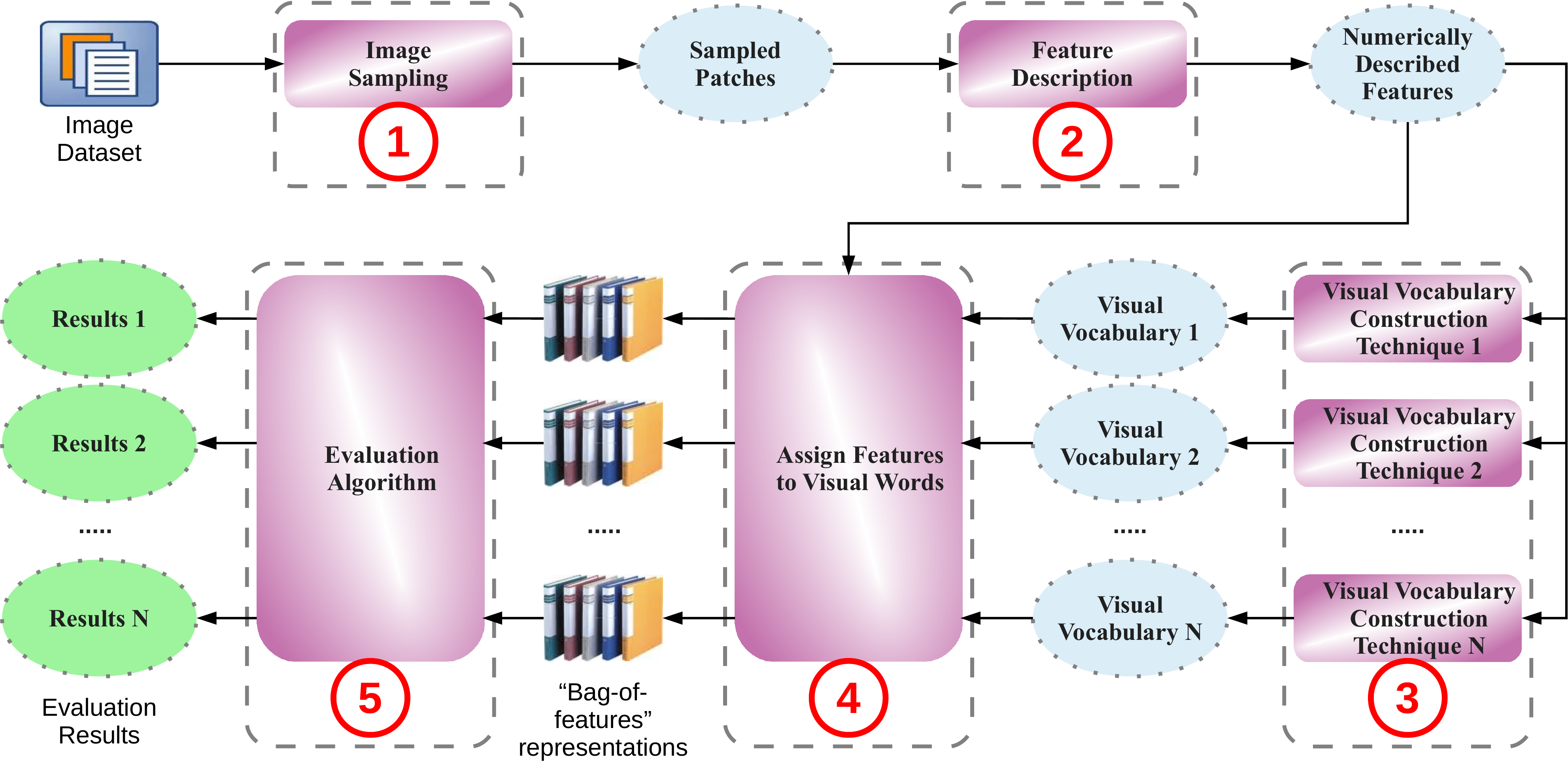}
	
	\caption{Schema for evaluating multiple visual vocabulary construction techniques.}
	\label{fig:evaluation-protocol}
\end{figure}

More precisely, given the fact that we perform the semantic injection at the level of the visual vocabulary construction, the experimental protocol streamlined in Figure~\ref{fig:evaluation-protocol} and further detailed in Section~\ref{subsec:experiments-protocol},  is designed to quantify the differences of performance due only to the visual vocabulary construction.
The evaluation is a five phase process, out of which four phases (\textit{1, 2, 3} and \textit{5}) are identical for all techniques.
Phases \textit{1} to \textit{4} correspond to the \emph{BoF} representation construction (see Figure~\ref{fig:bag-of-features}), while the last phase corresponds to the learning algorithm.

We summarize here after each of these phases, which are further detailed in the next sections:
\begin{itemize}
	\item \textit{phase~1}: \textbf{image sampling}, identical for all compared approaches;
	\item \textit{phase~2}: \textbf{feature numerical description} of patches, identical for all compared approaches;
	\item \textit{phase~3}: \textbf{visual vocabulary construction}, using the baseline approaches and our semantically-enriching approaches;
	\item \textit{phase~4}: \textbf{feature assignment to visual words}, identical for all compared approaches;
	\item \textit{phase~5}: \textbf{learning algorithm}, each resulted representation is used with two classifiers (a clustering-based and an SVM), identical for all compared approaches.
\end{itemize}

\subsection{Experimental protocol}
\label{subsec:experiments-protocol}

Starting from a given image dataset, we construct, for each image, four \emph{BoF} representations corresponding to the four evaluated visual vocabulary construction techniques (in \textit{phase~3}).
The image sampling (\textit{phase~1}), the feature description (\textit{phase~2}) and the image description (\textit{phase~4}) are performed each time using the same algorithms and with the same parameters.
In the end, the performances of each obtained \emph{BoF} representation are measured and compared in the context of a content-based image classification task (detailed in Section~\ref{subsec:experiments-learning-task}).
The visual vocabulary construction phase is the only phase to vary between the different constructed representations.
Therefore, we consider the classifier performance differences a direct consequence of the vocabulary construction.

\paragraph{The invariant \textit{phases 1, 2} and \textit{4}}
In \textit{phase~1}, images are sampled using a Hessian-Affine region detector and patches are described, in \textit{phase~2}, using the SIFT descriptor~\cite{LOW04}.
We use the default parameters for these algorithms and we keep them unchanged during the experiments.
The visual vocabulary is constructed in \textit{phase~3} using the construction technique to be evaluated.
In \textit{phase~4}, the final numerical representation is created, for each image, by associating features to visual words, using the $tf$ term weighting scheme.
To reduce the hazard component that appears in all the considered techniques, each construction is repeated 3 times and average results are presented.

\paragraph{Compared vocabulary construction techniques (\textit{phase~3})}

Four visual vocabulary construction techniques are evaluated: two classical techniques \textbf{random}, \textbf{random+km} and our proposals \textbf{model} and \textbf{filt+model}.
\textbf{random} constructs a random vocabulary (features are randomly chosen to serve as visual words).
For \textbf{random+km}, we take the random features selected previously and we ameliorate them by using the \textbf{ameliorate\_using\_\KMeans{}} function presented in Section~\ref{subsec:proposals-label-based-model-generation}.
\textbf{random+km} is the baseline construction technique presented in Section~\ref{subsec:introduction-baseline-approach}.
\textbf{model} is our proposal for dedicated vocabulary construction presented in Algorithm~\ref{algo:dedicated-visual-vocabulary-creationg}.
In \textbf{filt+model} we applied the filtering technique presented in Algorithm~\ref{algo:point-of-interest-filtering-algorithm} as a pre-processing phase before the dedicated vocabulary construction.

\subsection{The learning task: content-based image classification}
\label{subsec:experiments-learning-task}

Each of the image representations obtained as shown in the previous sections, are used in a content-based image classification task.
Two classifiers, an SVM and a clustering-based classifier, are trained and evaluated on each representation, as described in the following paragraphs.
The SVM classifier is chosen since it is the most widely used classifier in the object-based image classification literature, showing some of the best results and being a \textit{de facto} standard.
The \KMeans{}-based classifier is used in order to asses our proposals with a second, weaker classifier.
The choice of these classifiers is not unique and any other classifier can be used.
In the following sections, we evaluate our proposals by measuring the performance gain for any given classifier. 

\paragraph{The SVM classifier~\cite{COR95}}
The SVM classifier evaluation respects the experimental setup recommended by the authors of the \texttt{Caltech101}\footnote{\url{http://www.vision.caltech.edu/Image_Datasets/Caltech101/}} dataset.
We used the SVM implementation present in the LibSVM\cite{CHA11} library, using a linear kernel and default parameter values.
One of the challenges when evaluating in Data Mining is the disequilibrium between the class cardinality (usually it is the minority class that is of interest).
This disequilibrium can cause errors in estimating the generalization error of the constructed model.
Usually, the disequilibrium is the result of a certain reality in the population from which the sample was extracted (\textit{e.g.} the population of sick individuals is a minority compared to the healthy population).
But in the case of image datasets like \texttt{Caltech101}, the disequilibrium is only the result of the choice of its creator and represents no reality that needs to be taken into account.
We choose to equilibrate the classes before training the classifier, by randomly selecting 30 examples for each label to be part of the learning set.
15 images in the learning corpus are randomly selected to be part of the labeled set $\mathcal{P}_1$.
We test on all remaining individuals, which means that the generalization error on majority classes will be better estimated.
Evaluation indicators are calculated for each class and we report only the non-weighted averages.
The process is repeated 10 times: we create 10 learning sets and the corresponding 10 testing sets.
We report the average performances over the 10 executions.
The results are expressed using the True Positive Rate, because this measure is usually used in the literature when reporting results on \texttt{Caltech101} and \texttt{RandCaltech101}.

\paragraph{A clustering-based classifier}
The clustering-based evaluation task is inspired from the unsupervised information retrieval field and it is based on clustering.
A learning set of the image collection is clustered into a number of clusters and each cluster is assigned a label, using a majority vote.
Each image in the test corpus is assigned to its nearest centroid and it is given the predicted label of the cluster.
Predicted labels are compared to the real labels and classical information retrieval measures (\textit{i.e.}, precision, recall, \FMeasure{}) are calculated.

The evaluation of the clustering-based classifier is performed using a stratified holdout strategy.
The images are divided into a learning corpus (67\% of images in each category) and a test corpus (33\% of the images in each category).
50\% of images in the learning corpus are randomly selected to be part of the labeled set $\mathcal{P}_1$.
For the rest, the labels are hidden.
Images in the learning set are then clustered into $nc$ clusters using \KMeans{}.
$nc$ varies between 50 and 1000 (step 50) for \texttt{Caltech101} and \texttt{RandCaltech101} and between 3 and 90 (step 3) for \texttt{Caltech101-3} (\texttt{Caltech101-3} contains only 3 classes, see Section~\ref{subsec:experiments-datasets}).
To eliminate the effect of disequilibrium between class sizes, we calculate and report the non-weighted averages over tags of these indicators.
To measure the classification accuracy, we use the \FMeasure{} (the harmonic average of precision and recall), a classical Information Retrieval measure.
For each combination (vocabulary dimension, $nc$, vocabulary algorithm), the clustering and prevision phase is repeated 25 times, to eliminate the influence of the random initialization of the \KMeans{} in the clustering-based classifier.

\subsection{Datasets}
\label{subsec:experiments-datasets}

Experiments were performed on the \texttt{Caltech101}~\cite{FEI07} and \texttt{RandCaltech101}~\cite{KIN10} datasets.
\texttt{Caltech101} contains 9144 images, most of them in medium resolution (300$\times$300 pixels).
It is a heterogeneous dataset, having 101 object categories and one reserve.
Each category class is considered to be a label.
Spatial positioning of objects is not used, therefore positioning ourselves in a weakly supervised context.
Some authors argue that \texttt{Caltech101} is not diverse enough and that backgrounds often provide more information than the objects themselves.
\texttt{RandCaltech101} is obtained from \texttt{Caltech101} by randomly modifying the backgrounds and the posture (position, orientation) of objects.
It has been shown~\cite{KIN10} that classification is more challenging on \texttt{RandCaltech101} than on \texttt{Caltech101}.

Because \texttt{Caltech101} is an unbalanced dataset, with category sizes ranging from 31 to 800 images, we have taken 3 out of the biggest categories (\textit{airplanes}, \textit{Motorbikes} and \textit{Faces\_easy}) and created another corpus, denoted \texttt{Caltech101-3}.
It contains 2033 images.
The advantage of the new corpus is that it provides many examples for each category and it is balanced category-wise.
This allows us to study how our propositions behave on both balanced and unbalanced datasets.

\subsection{Qualitative evaluation}
\label{subsec:experiments-qualitative-evaluation}

\begin{figure}[h]
  \centering
  	\subfloat{\includegraphics[height=0.08\textheight]{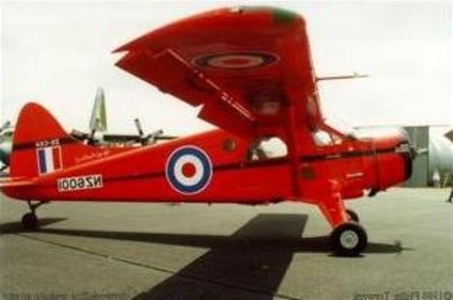}}  
  	\subfloat{\includegraphics[height=0.08\textheight]{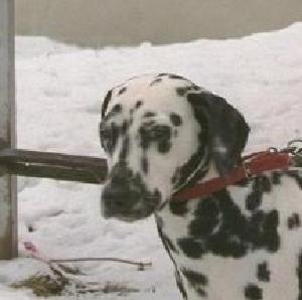}} 
  	\subfloat{\includegraphics[height=0.08\textheight]{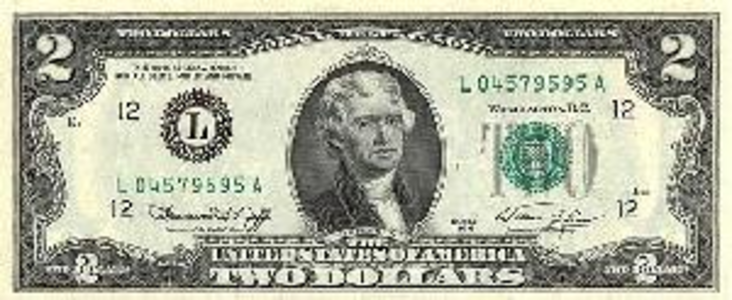}} 
  	\subfloat{\includegraphics[height=0.08\textheight]{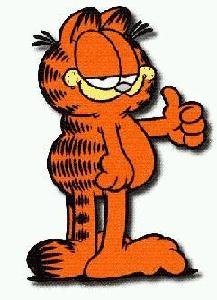}} 
  	\subfloat{\includegraphics[height=0.08\textheight]{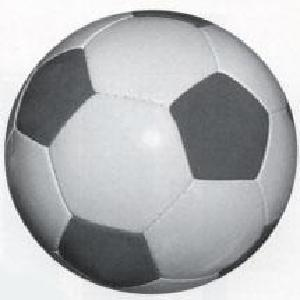}} 
  	\subfloat{\includegraphics[height=0.08\textheight]{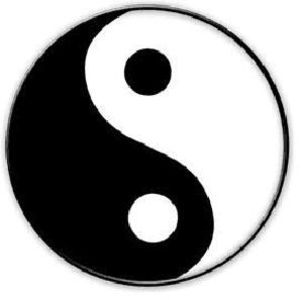}}
  		\hfill  	
  	\subfloat{\includegraphics[height=0.08\textheight]{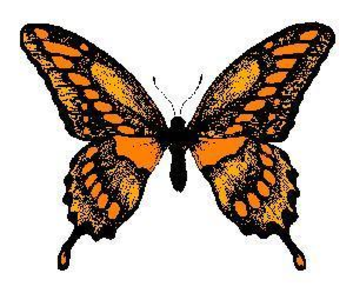}}  
  	\subfloat{\includegraphics[height=0.08\textheight]{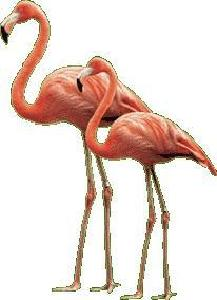}} 
  	\subfloat{\includegraphics[height=0.08\textheight]{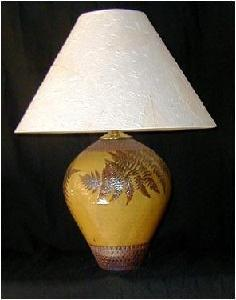}} 
  	\subfloat{\includegraphics[height=0.08\textheight]{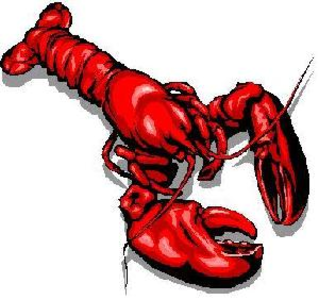}}   	
  	\subfloat{\includegraphics[height=0.08\textheight]{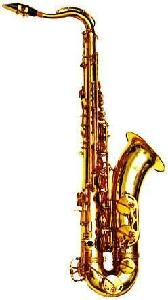}} 
  	\subfloat{\includegraphics[height=0.08\textheight]{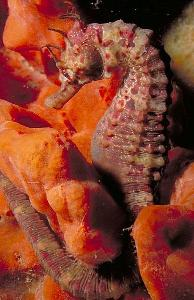}} 
  	\subfloat{\includegraphics[height=0.08\textheight]{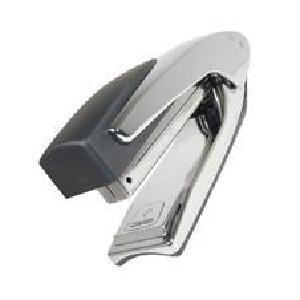}} 
  	\subfloat{\includegraphics[height=0.08\textheight]{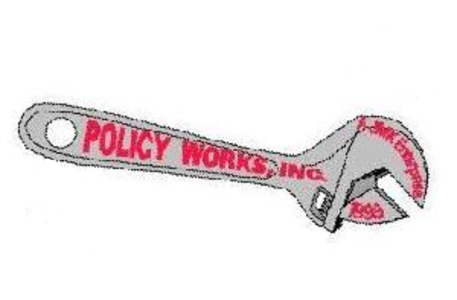}} 
  
	\caption{Example of images from ``easy'' classes (top row) and ``difficult'' classes (bottom row)}
  	\label{fig:tags-examples}
\end{figure}

In a classification tasks, some classes are naturally easier to recognize than others.
This happens when the numerical description is better adapted to translate them into a separable numerical space.
On \texttt{Caltech101}, the best classification scores are almost invariably obtained by the same categories, independent of the choice of visual construction algorithms or parameters.

Figure~\ref{fig:tags-examples} shows some examples of images belonging to ``easy classes'', categories that obtain good classification scores (on the upper row), and examples of ``difficult classes'', categories that obtain low scores (on the bottom row).
The objects belonging to the ``easy classes'' either appear in the same posture in all examples or they have a specific color pattern that makes them easily recognisable.
Most of the examples of \textit{airplanes} and \textit{garfield} appear with the same shape, size and orientation.
Other categories like \textit{yin\_yang}, \textit{soccer\_ball} or \textit{dalmatian} have a specific white-black alternation pattern, which makes them easily recognizable even in the real world.
By contrast, the objects depicted in picture of ``difficult classes'', like \textit{seahorse} or \textit{butterfly} appear in different colors, multiple postures and sometimes hidden in the background.

\begin{table}[!t]
\caption{``Easy'' classes and ``difficult'' classes in \texttt{Caltech101} and \texttt{RandCaltech101}}
\label{tab:easy-difficult-classes}
\centering

\begin{tabular}{@{}cc|cc@{}} \toprule
\multicolumn{ 2}{c|}{\textbf{``Easy'' classes}} & \multicolumn{ 2}{c}{\textbf{``Difficult'' classes}} \\ 

\textit{Caltech101} & \textit{RandCaltech101} & \textit{Caltech101} & \textit{RandCaltech101} \\ \midrule

\textit{airplanes} & \textbf{accordion} & \textbf{beaver} & \textbf{bass} \\ 

\textit{car\_side} & \textit{airplanes} & \textit{buddha} & \textbf{binocular} \\ 

\textit{dalmatian} & \textit{car\_side} & \textit{butterfly} & \textbf{brontosaurus} \\
 
\textit{dollar\_bill} & \textit{dalmatian} & \textbf{ceiling\_fan} & \textit{buddha} \\ 

\textit{Faces\_easy} & \textit{dollar\_bill} & \textbf{cougar\_body} & \textit{butterfly} \\ 

\textit{garfield} & \textit{Faces\_easy} & \textit{crab} & \textit{crab} \\ 

\textbf{grand\_piano} & \textit{garfield} & \textit{crayfish} & \textit{crayfish} \\ 

\textbf{Leopards} & \textbf{laptop} & \textit{cup} & \textbf{crocodile} \\ 

\textbf{metronome} & \textit{Motorbikes} & \textit{dragonfly} & \textit{cup} \\ 

\textit{Motorbikes} & \textit{panda} & \textit{ewer} & \textit{dragonfly} \\ 

\textit{panda} & \textit{snoopy} & \textbf{ferry} & \textit{ewer} \\ 
\textbf{scissors} & \textit{soccer\_ball} & \textit{flamingo} & \textit{flamingo} \\ 

\textit{snoopy} & \textit{stop\_sign} & \textit{flamingo\_head} & \textit{flamingo\_head} \\
 
\textit{soccer\_ball} & \textit{watch} & \textit{ibis} & \textbf{gerenuk} \\ 

\textit{stop\_sign} & \textit{windsor\_chair} & \textit{kangaroo} & \textbf{helicopter} \\ 

\textbf{tick} & \textit{yin\_yang} & \textit{lamp} & \textit{ibis} \\ 

\textit{watch} &  & \textit{lobster} & \textit{kangaroo} \\ 

\textit{windsor\_chair} &  & \textit{mandolin} & \textit{lamp} \\ 

\textit{yin\_yang} &  & \textit{mayfly} & \textit{lobster} \\ 

 &  & \textit{minaret} & \textit{mandolin} \\ 
 &  & \textit{pigeon} & \textit{mayfly} \\ 
 &  & \textit{platypus} & \textbf{metronome} \\ 
 &  & \textbf{pyramid} & \textit{minaret} \\ 
 &  & \textbf{rhino} & \textbf{okapi} \\ 
 &  & \textit{saxophone} & \textit{pigeon} \\
 &  & \textbf{schooner} & \textit{platypus} \\ 
 &  & \textit{sea\_horse} & \textit{saxophone} \\ 
 &  & \textit{stapler} & \textit{sea\_horse} \\ 
 &  & \textbf{strawberry} &  \textit{stapler} \\ 
 &  & \textbf{wild\_cat} &  \textit{wrench} \\
 &  & \textit{wrench} &  \\ \bottomrule
\end{tabular}

\end{table}

We perform the same analysis on \texttt{RandCaltech101}.
Table~\ref{tab:easy-difficult-classes} presents a comparative view of ``easy classes'' and ``difficult classes'' constructed for \texttt{Caltech101} and \texttt{RandCaltech101}, with the non-identical categories (between the two datasets) printed in boldface.
We observe the high degree of overlapping of the constructed sets: most of the ``easy classes'' in \texttt{Caltech101} also appear as ``easily'' recognizable for \texttt{RandCaltech101}.
Similarly, difficult classes on \texttt{Caltech101} remain difficult on \texttt{RandCaltech101}.
In Table~\ref{tab:easy-difficult-classes}, 
the only category that changes difficulty is \textit{metronome}, which is an ``easy class'' in \texttt{Caltech101} and a ``difficult class'' in \texttt{RandCaltech101}.
This proves that the background randomization performed in order to create \texttt{RandCaltech101}, while it makes the dataset more challenging to classify as a whole, does not change the relative difficulty between categories.
Categories that obtain good classification scores for \texttt{Caltech101} also obtain good scores for \texttt{RandCaltech101}.

\subsection{Quantitative evaluation}
\label{subsec:experiments-quantitative-evaluation}

In this section, we show how the performances of the two classifiers vary, depending on the visual vocabulary construction technique and the size of the visual vocabulary.
We show that the semantically-enriched representation clearly outperform the baseline approach, mostly by increasing the score of ``difficult'' categories, and we discuss the overfitting.
For all the experiments presented in this subsection, the parameter $\alpha$ (introduced in Equation~\ref{eq:threshold-alpha}) of the filtering heuristic \textbf{filt+model} is set at one ($\alpha = 1$) and its influence is studied later, in Section~\ref{subsec:experiments-parameter-alpha}.

\begin{figure}[t]
	\centering
	
	\subfloat[]{
		\label{subfig:typical-caltech101}
		\includegraphics[width=0.47\textwidth]{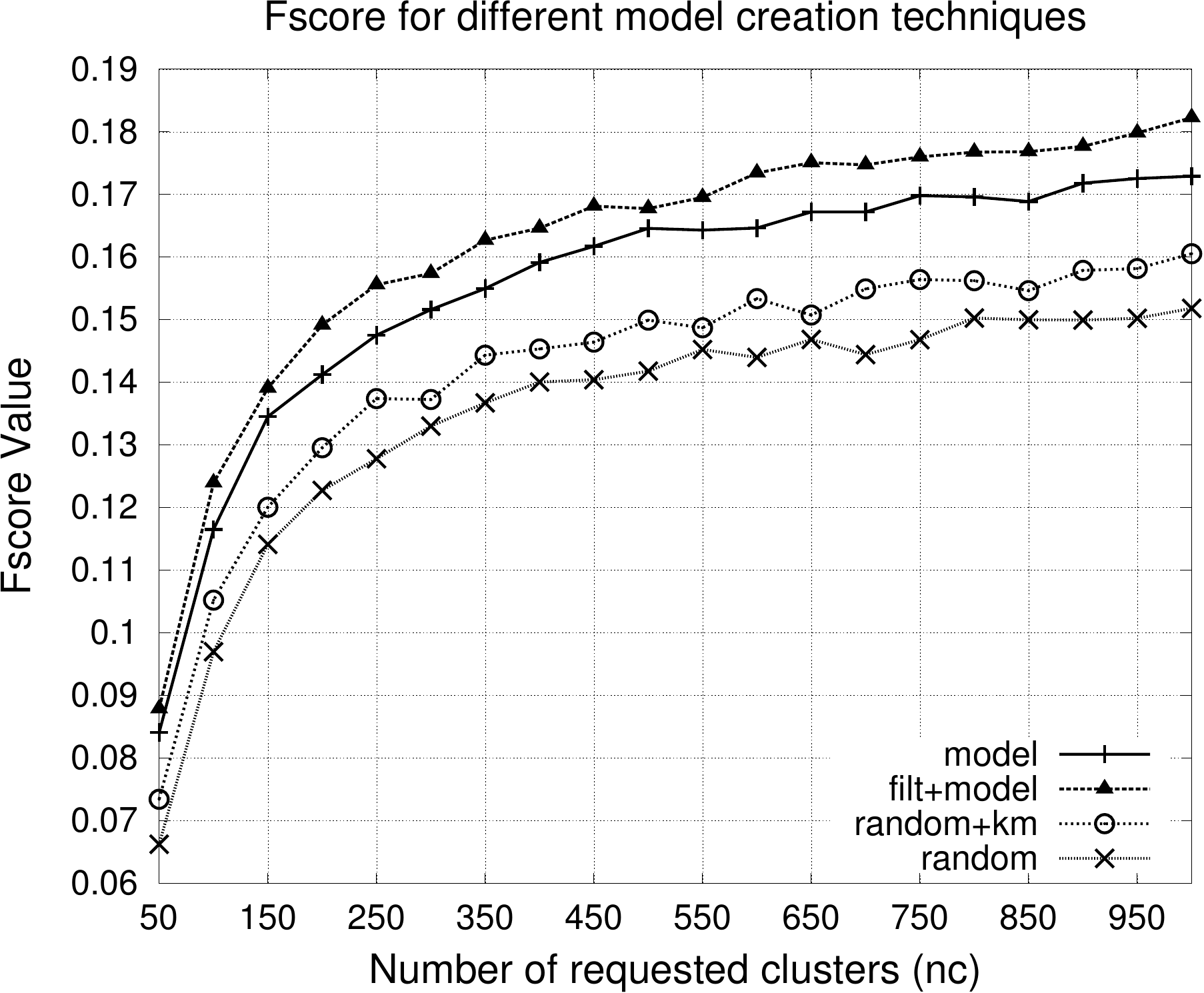}
	}
	\hfill
	\subfloat[]{
		\label{subfig:typical-randcaltech101}
		\includegraphics[width=0.47\textwidth]{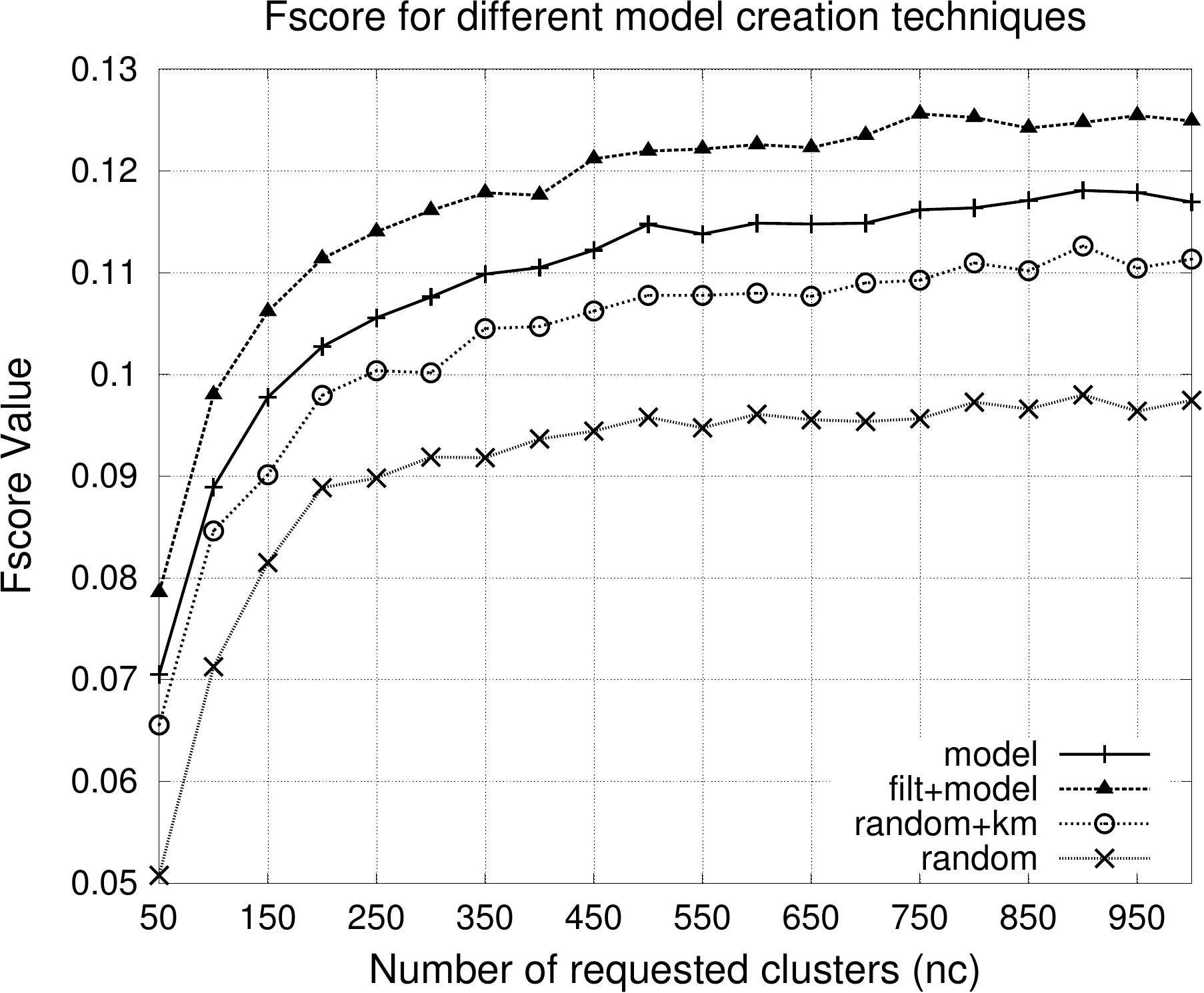}
	}
	
	\caption{A typical \FMeasure{} evolution for the clustering-based classifier for $m = 1000$ on \texttt{Caltech101} (a) and on \texttt{RandCaltech101} (b)}
	\label{fig:typical-prediction}
\end{figure}

\paragraph{Aggregating the number of clusters in the \textit{clustering-based classifier}}

When using the clustering-based classification algorithm, for a fixed visual vocabulary size, varying the number of clusters $nc$ leads to an \FMeasure{} variation as shown in Figure~\ref{fig:typical-prediction}.
For all visual vocabulary techniques, the \FMeasure{} has a steep amelioration for lower values of $nc$ and stabilizes once $nc$ reaches a value which is approximately two-three times bigger than the number of categories.
Starting from this point \FMeasure{} augments slowly and reaches its theoretical maximum when $nc$ equals the number of individuals in the testing set.
Due to the fact that once stabilized, the score can be considered relatively constant, we compute the mean \FMeasure{} over all the values for $nc$.
We obtain, for each visual vocabulary dimension, an aggregated \FMeasure{}.

\begin{figure}[!t]
	\centering
	
	\subfloat[]{ 
		\label{subfig:caltech101-prediction}
		\includegraphics[width=0.48\textwidth]{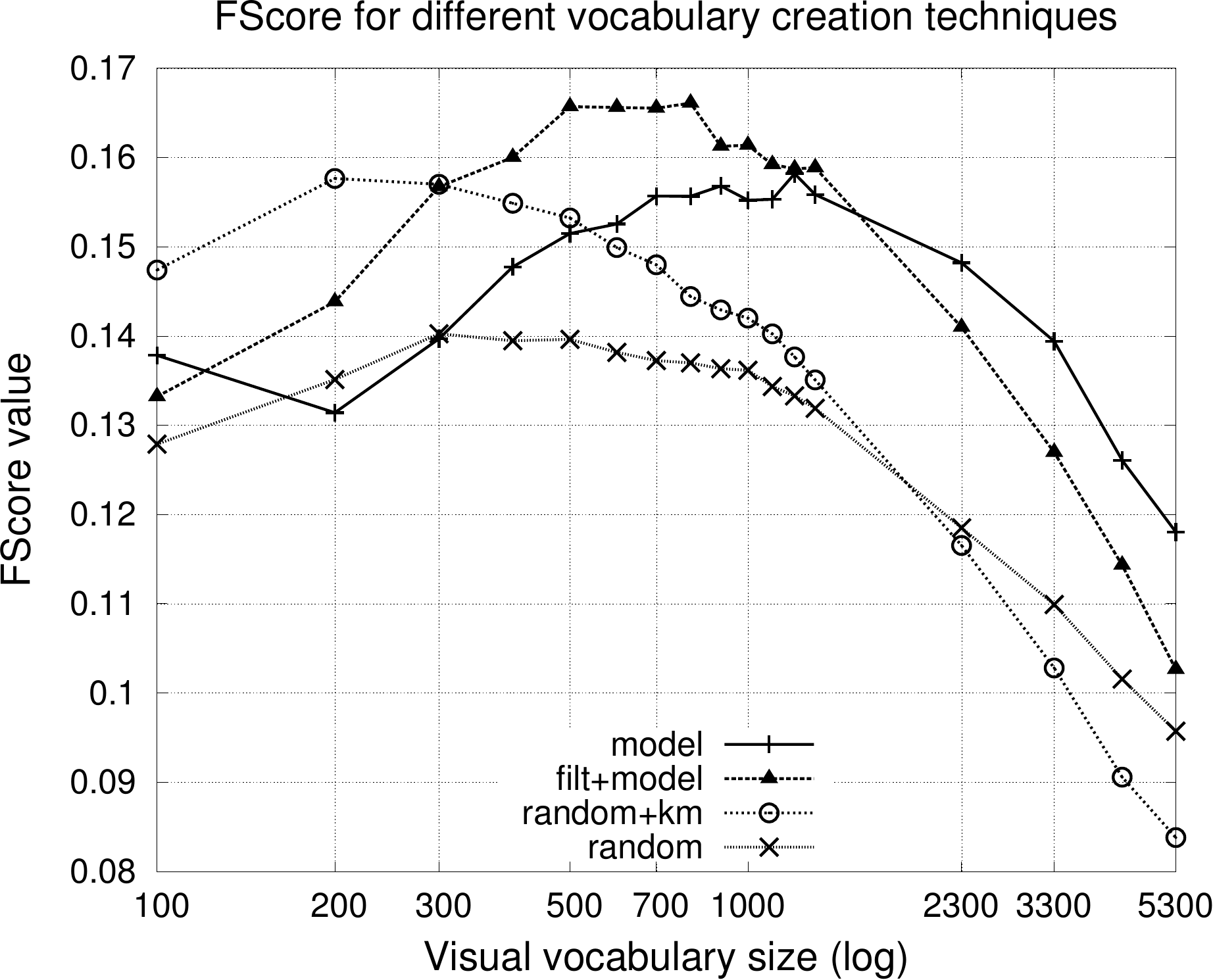} 
	}
	\hfill
	\subfloat[]{ 
		\label{subfig:caltech101-svm}
		\includegraphics[width=0.48\textwidth]{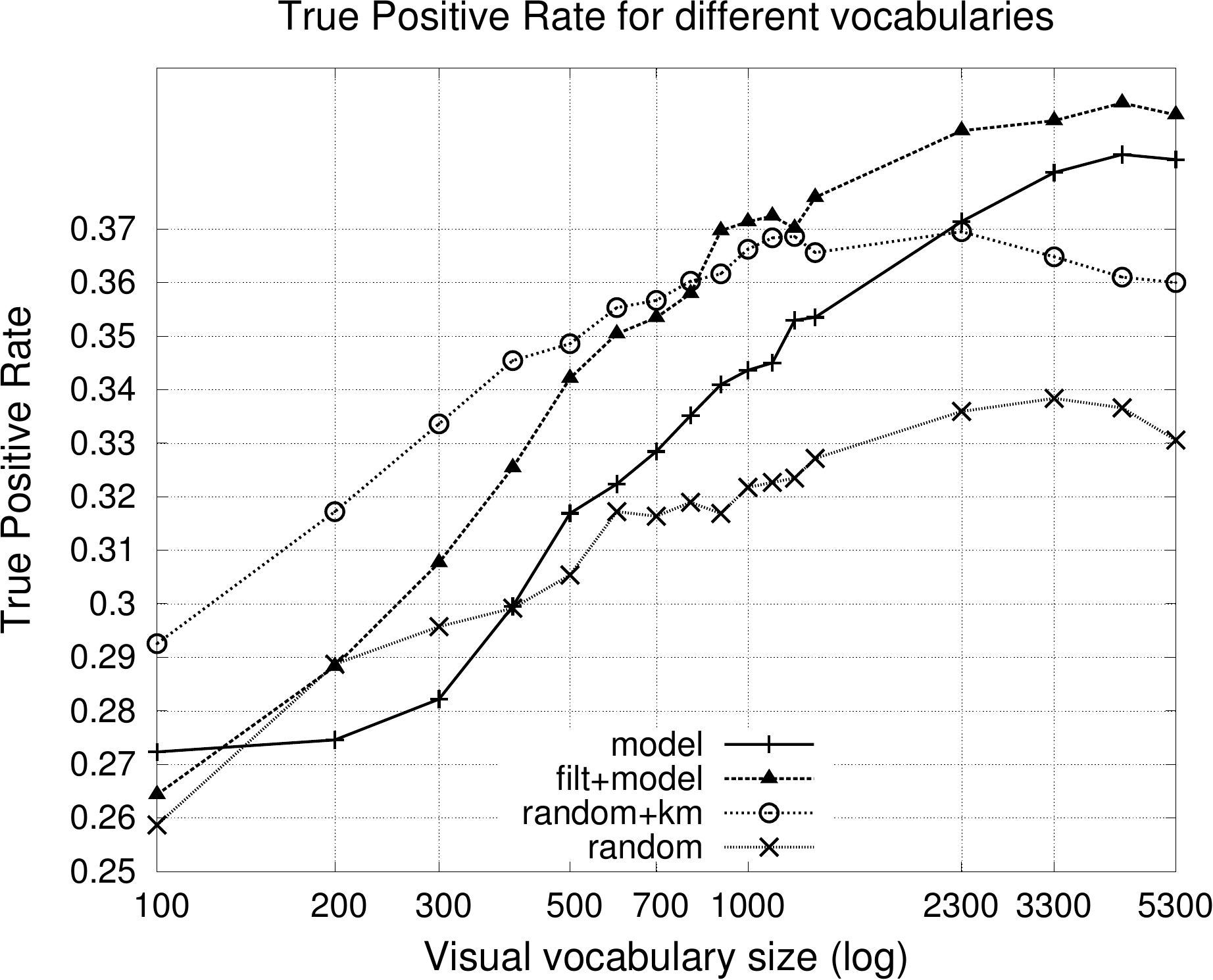} 
	}
	
	\caption{\texttt{Caltech101}: Aggregated \FMeasure{} with clustering-based classifier (a) and $True Posivite Rate$ for SVM (b) as functions of the vocabulary size}
	\label{fig:caltech101}
\end{figure}

\begin{figure}[!t]
	\centering

	\subfloat[]{ 
		\label{subfig:caltech101-3-prediction}
		\includegraphics[width=0.48\textwidth]{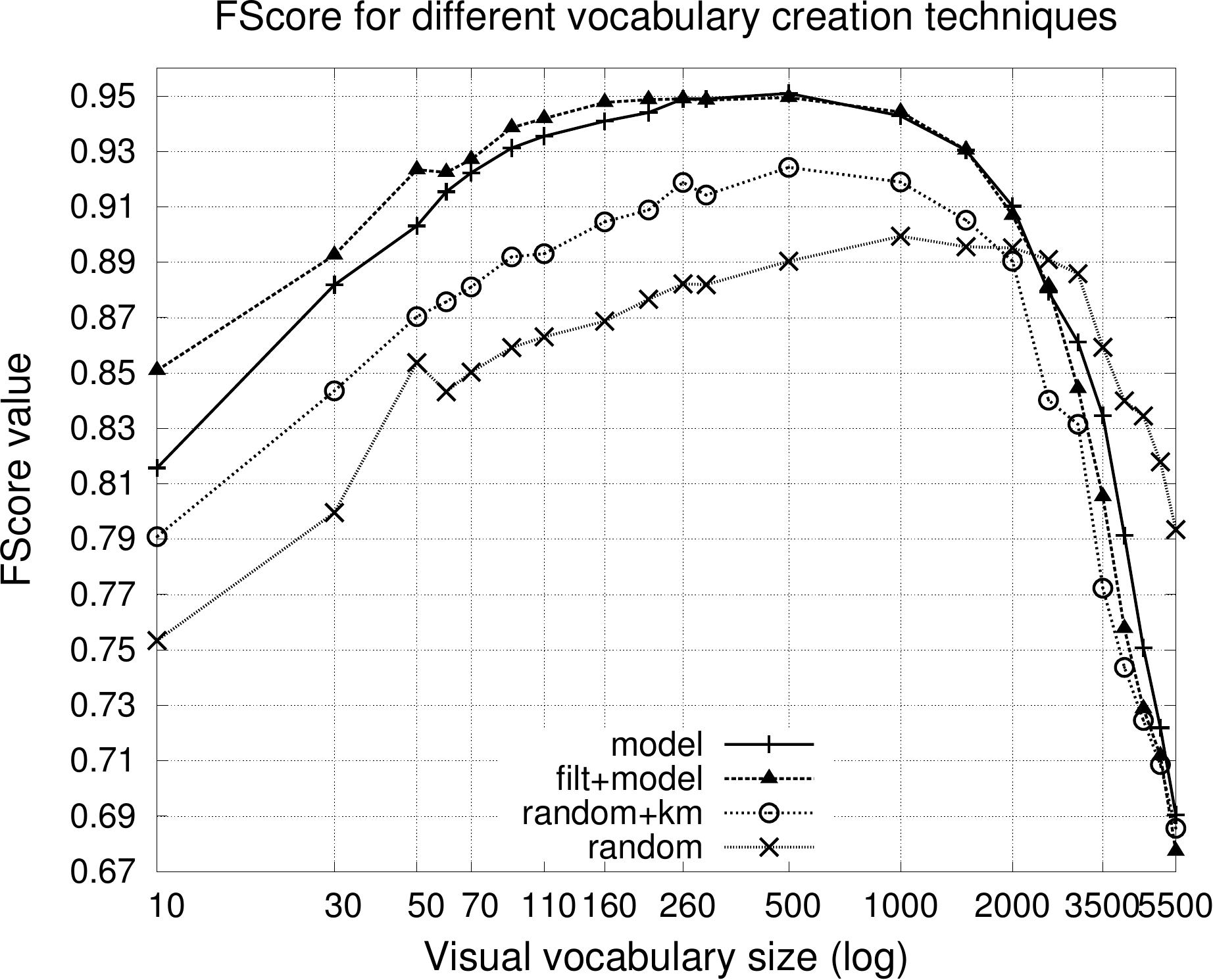} 
	}
	\hfill
	\subfloat[]{ 
		\label{subfig:caltech101-3-svm}
		\includegraphics[width=0.48\textwidth]{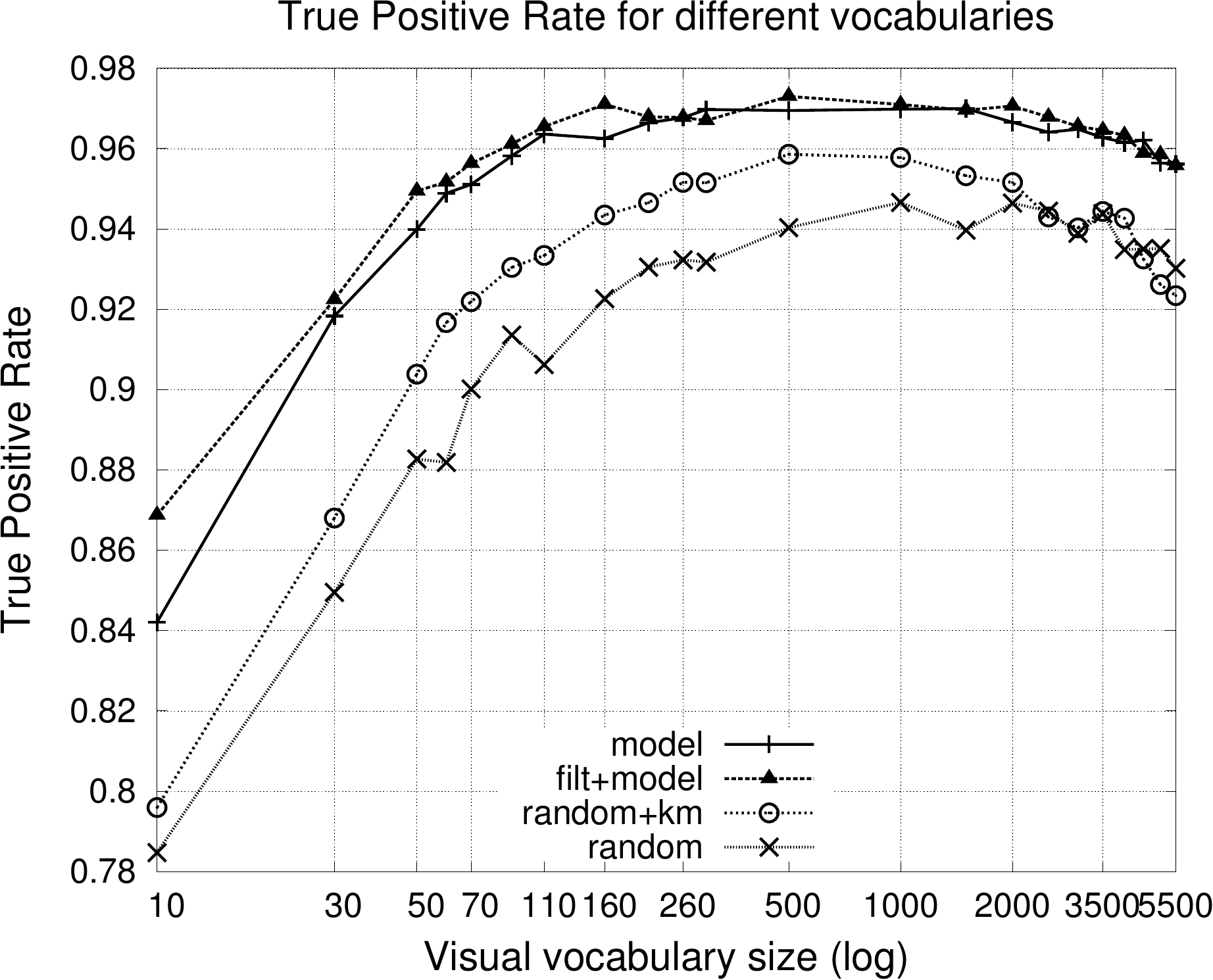}
	}
	 	
	\caption{\texttt{Caltech101-3}: Aggregated \FMeasure{} with clustering-based classifier (a) and $True Posivite Rate$ for SVM (b) as functions of the vocabulary size}
	\label{fig:caltech101-3}
\end{figure}

\begin{figure}[!t]
	\centering
	
	\subfloat[]{
		\label{subfig:randcaltech101-prediction}		
		\includegraphics[width=0.48\textwidth]{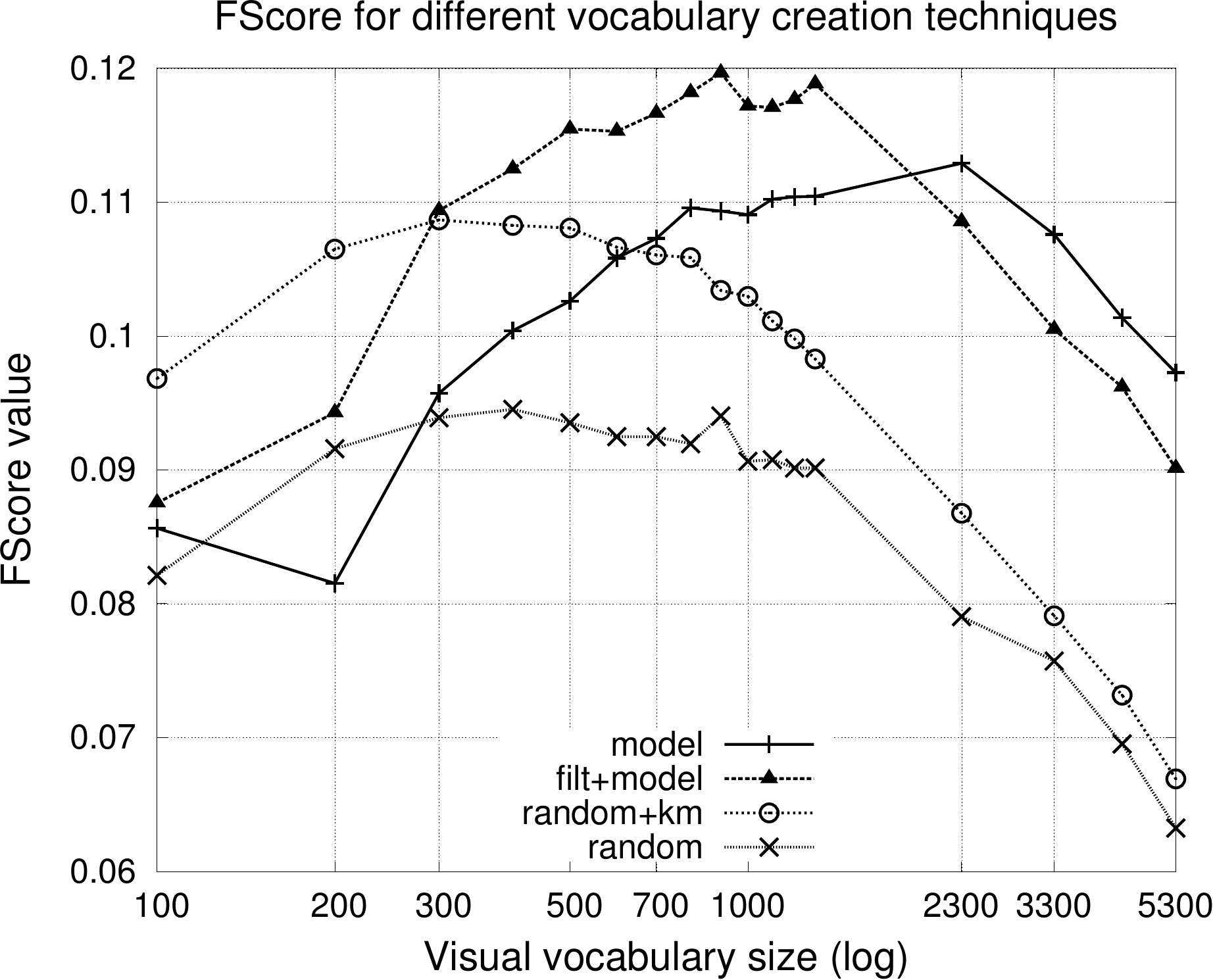}
	}
	\hfill
	\subfloat[]{ 
		\label{subfig:randcaltech101-svm}
		\includegraphics[width=0.48\textwidth]{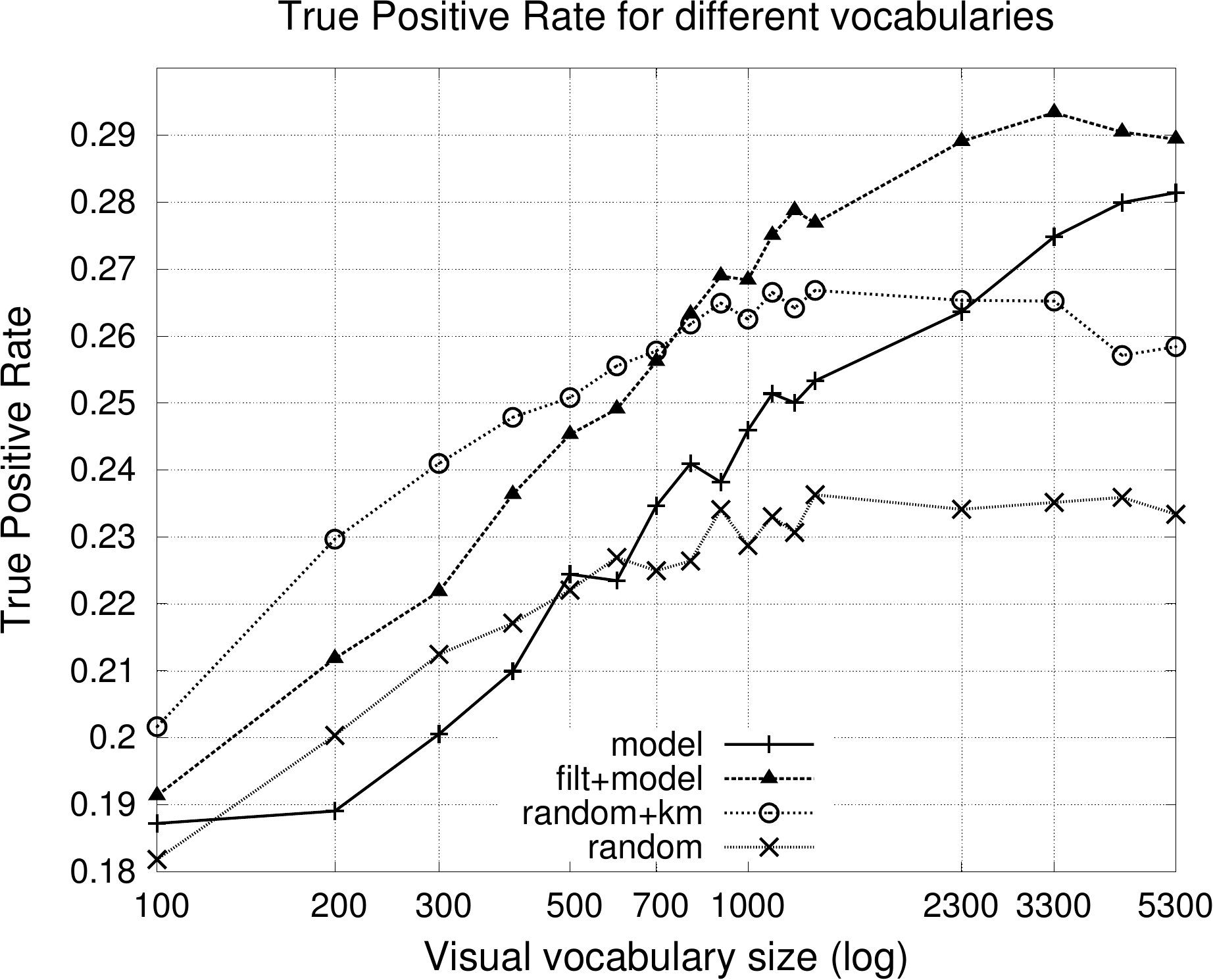} 
	}
	
	\caption{\texttt{RandCaltech101}: Aggregated \FMeasure{} with clustering-based classifier (a) and $True Posivite Rate$ for SVM (b) as functions of the vocabulary size}
	\label{fig:randcaltech101}
\end{figure}

\paragraph{Obtained graphics}
Figures~\ref{fig:caltech101},~\ref{fig:caltech101-3} and~\ref{fig:randcaltech101} present the score evolution as a function of the visual vocabulary size on, respectively, the datasets \texttt{Caltech101}, \texttt{Caltech101-3} and \texttt{RandCaltech101}.
More precisely, Figures~\ref{subfig:caltech101-prediction},~\ref{subfig:caltech101-3-prediction} and~\ref{subfig:randcaltech101-prediction} show the evolution of the aggregated \FMeasure{}, for the \textit{clustering-based classifier}, and Figures~\ref{subfig:caltech101-svm},~\ref{subfig:caltech101-3-svm} and~\ref{subfig:randcaltech101-svm} show the variation of the $True Positive Rate$, using the SVM classifier.

We make vary the vocabulary dimension between 100 and 5300 for \texttt{Caltech101} and \texttt{RandCaltech101} and between 10 and 5500 for the \texttt{Caltech101-3}, using a variable step.
For the three datasets, the horizontal axis is logarithmic.
When observing the graphics for every tuple \emph{(dataset, classifier, vocabulary construction technique)}, we observe the pattern of a dome-like shape, corresponding to the three phases: under-fitting, maximum performance and overfitting.
We analyze more in detail the overfitting behavior for each vocabulary construction technique in Section~\ref{subsec:experiments-overfitting}.
Furthermore, the somehow low results obtained by the clustering-based classifier can be explained by the fact that the clustering-based classifier is a weak classifier (\textit{i.e.}, a classifier which perform only slightly better than a random classifier), whereas the SVM is a strong classifier.

\paragraph{Results interpretation}
When comparing the relative performances of the different techniques presented in Figures~\ref{fig:caltech101},~\ref{fig:caltech101-3} and~\ref{fig:randcaltech101}, we observe that our semantic-aware proposals (\textit{i.e.}, \textbf{model} and \textbf{filt+model}) generally obtain better results than the generic (\textbf{random+km}) and \textbf{random} ones.
The three regions of evolution are wider (they enter overfitting later) for \textbf{model} and \textbf{filt+model} than for \textbf{random} and \textbf{random+km}.
On the other hand, they also exit the under-fitting later.
The generic \textbf{random+km} obtains better results than \textbf{model} and \textbf{filt+model}, for lower dimensions of visual vocabulary, on \texttt{Caltech101} and \texttt{RandCaltech101}.
After exiting the under-fitting region, \textbf{model} and \textbf{filt+model} constantly obtain better scores than \textbf{random+km}, even when overfitted.
Applying our filtering proposal (\textbf{filt+model}) consistently provides a plus of performance (over \textbf{model}), but also causes the visual vocabulary to enter overfitting earlier.

\begin{table}[htb]
\centering

\caption{Average gain of performance relative to \textbf{random}.}
\label{tab:gain}

\begin{tabular}{llrrr}
\toprule
 & \multicolumn{1}{l}{} & \multicolumn{1}{c}{\textbf{model}} & \multicolumn{1}{c}{\textbf{filt+model}} & \multicolumn{1}{c}{\textbf{random+km}} \\ \midrule
 
\multirow{3}{*}{\begin{sideways}pred.\end{sideways}} 
& \texttt{Caltech101} & 13.96\% & \textbf{15,69\%} & 4,36\% \\ 
& \texttt{Caltech101-3} & 6.58\% & \textbf{7.36\%} & 2.73\% \\ 
& \texttt{RandCaltech101} & 20,49\% & \textbf{26,27\%} & 12,07\% \\   \cmidrule{ 2 - 5}

\multirow{3}{*}{\begin{sideways}SVM\end{sideways}} 
& \texttt{Caltech101} & 5,98\% & 12,02\% & \textbf{12,05\%} \\  
& \texttt{Caltech101-3} & 4,71\% & \textbf{5.24\%} & 1,90\% \\  
& \texttt{RandCaltech101} & 5,89\% & \textbf{15,20\%} & 13,21\% \\  \bottomrule
\end{tabular}
\end{table}

Table \ref{tab:gain} gives the average gain of performance relative to \textbf{random} for the generic \textbf{random+km} and our semantic-aware proposals \textbf{model} and \textbf{filt+model}.
For the clustering-based classifier, we show the average relative \FMeasure{} gain, while for the SVM we show the average relative $True Positive Rate$ gain.
The best scores for each dataset are shown in bold.
In five out of six cases, the best scores are obtained by \textbf{filt+model}.
\textbf{model} also performs better than the generic \textbf{random+km} in four out of the six cases.
This shows that a semantically-enriched representation outperforms the generic method \textbf{random+km} in a classification task.
The maximum gain of performance is achieved on \texttt{RandCaltech101}, where, by eliminating the background noise, our filtering algorithm considerably improves the classification performances.
When used with the SVM classifier on \texttt{Caltech101} and \texttt{RandCaltech101}, the \textbf{model} technique obtains average scores lower than \textbf{random+km}.
This is because \textbf{model} exits the under-fitting later than the other techniques, thus lowering its average score (as shown in Figures~\ref{subfig:caltech101-svm} and~\ref{subfig:randcaltech101-svm}).

\begin{figure}[t]
	\centering
	
	\subfloat[]{ 
		\label{subfig:roc-caltech101}		
		\includegraphics[width=0.48\textwidth]{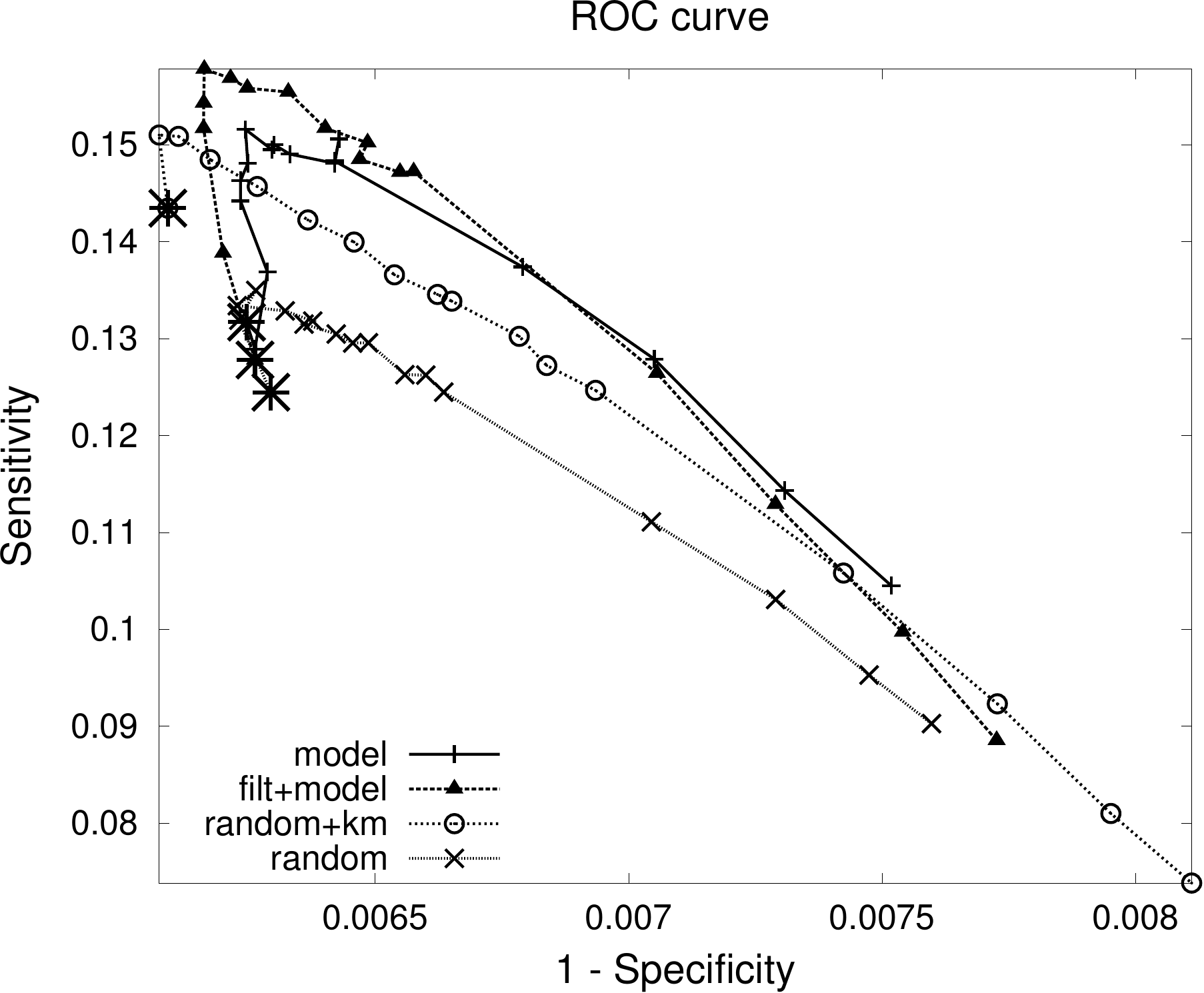} 
	}
	\hfill
	\subfloat[]{ 
		\label{subfig:roc-randcaltech101}
		\includegraphics[width=0.48\textwidth]{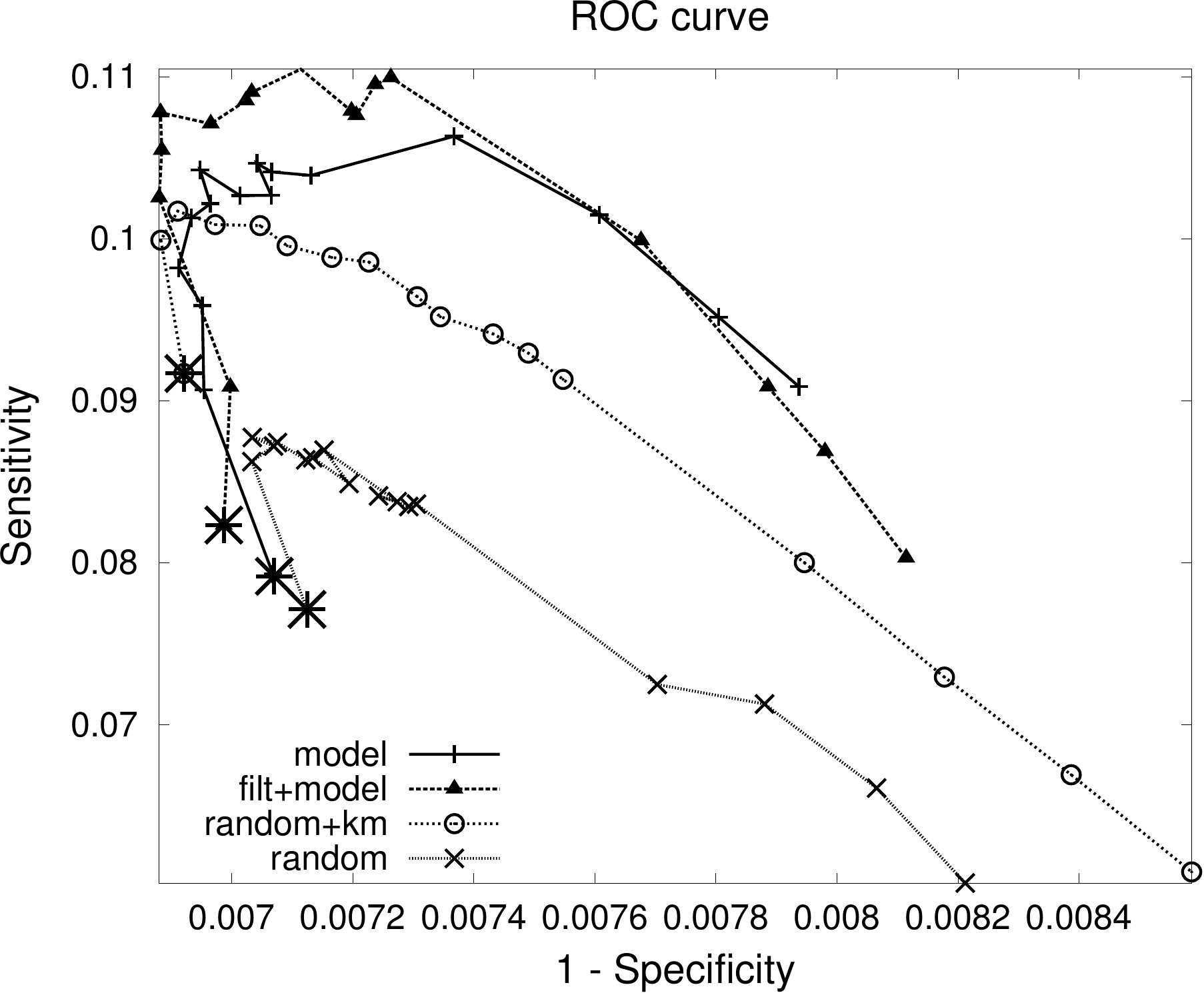} 
	}
	
	\caption{ROC curves: clustering-based classifier on \texttt{Caltech101} (a) and \texttt{RandCaltech101} (b)}
	\label{fig:roc-curves}
\end{figure}

\paragraph{The ROC curves}
Similar conclusions regarding the overfitting and the relative performances of the different visual vocabulary construction techniques can be drawn by plotting the evolution using \textit{ROC}~\cite{FAW06} curves.
Figure~\ref{fig:roc-curves} shows the ROC curves obtained using the clustering-based classifier on \texttt{Caltech101}(Figure~\ref{subfig:roc-caltech101}) and on \texttt{RandCaltech101} (Figure~\ref{subfig:roc-randcaltech101}).
The visual vocabulary size varied between 100 and 5300.
The sign \textbf{*} on the graphic indicates the smallest size.
The plots are zoomed to the relevant part.
Overfitting is clearly visible on the ROC curves.
All the curves start by climbing towards the ideal point $(0, 1)$ (first and second region on the graphics in Figures~\ref{subfig:caltech101-prediction} and ~\ref{subfig:randcaltech101-prediction}).
After reaching a maximum, the ROC curves start descending towards the ``worst'' point $(1,0)$, showing the overfitting region.
The curve corresponding to \textbf{filt+model} clearly dominates all the other, confirming the conclusions drawn from studying Table~\ref{tab:gain}: the proposed approaches and especially their combination in \textbf{filt+model}, achieve higher classification results.

\paragraph{Scores for ``easy'' and ``difficult'' categories}
In Section~\ref{subsec:experiments-qualitative-evaluation}, we have shown that in both \texttt{Caltech101} and \texttt{RandCaltech101} some classes are easier to learn than others.
Regardless of the visual vocabulary construction technique, ``easy classes'' obtain higher classification scores.
Nonetheless, the construction particularities of each technique influence the accuracy for difficult categories.
In \textbf{random}, features are randomly picked to serve as visual words.
Score differences between easy and difficult categories are pronounced and the overall accuracy is low.
The \KMeans{} iterations in \textbf{random+km} fit the visual vocabulary to ``easy'' classes.
Few categories achieve good scores, accentuating the gap between easy and difficult categories.
\textbf{model} and \textbf{filt+model} techniques achieve for ``difficult'' categories, better scores than \textbf{random} and \textbf{random+km}.
The visual vocabulary is representative for all categories and difficult categories like \textit{pyramid}, \textit{minaret} or \textit{stapler} obtain higher scores than those obtained with a baseline representation.

\subsection{Overfitting}
\label{subsec:experiments-overfitting}

\paragraph{Evaluating using the clustering-based classifier}
In the clustering-based classifier, for each pair (dataset, vocabulary construction technique), the \FMeasure{} graphic shows a dome-like shape with three regions.
In the first one, corresponding to low vocabulary dimensions, the visual vocabulary is under-fitted, there are not enough visual words to describe the objects~\cite{JIA07}.
Consequently, in the \textit{assign phase} (\textit{phase 4} in ``bag-of-features'' construction schema in Figure~\ref{fig:bag-of-features}), features are assigned to the same visual word even if they are not similar to each other.
The second region represents the interval in which the vocabulary obtains the best results.
In the third region (corresponding to large sizes of the visual vocabulary), performance degrades gradually.
This is due to the fact that, in the \textit{assign phase}, relevant features are grouped densely, while noise is evenly distributed.
Some of the visual words regroup relevant features, while other regroup only the noise.
As the visual vocabulary dimension augments, more and more visual words will regroup only noise.
This generates a numerical space of high dimensionality, which is separable only on a few dimension.
This leads to degrading the overall separability of the numerical space and the classification performances.

\paragraph{Evaluating using the SVM classifier}
The same conclusions apply for the SVM classifier.
Being a strong classifier, in Figures~\ref{subfig:caltech101-svm} (\texttt{Caltech101}) and~\ref{subfig:randcaltech101-svm} (\texttt{RandCaltech101}) the dome-shape is less visible for the SVM.
The overfitting appears for higher visual vocabulary sizes than in the clustering-based classifier.
For example, in Figure~\ref{subfig:randcaltech101-prediction}, for \textbf{random+km}, clustering-based classifier starts to overfit at a vocabulary size of 300.
When using the SVM, in Figure~\ref{subfig:randcaltech101-svm}, overfitting starts only at 1300.
The \textbf{model} technique does not appear to enter overfitting in Figure~\ref{subfig:randcaltech101-svm}.
But this is likely to happen for dimensions higher than 5300 (the maximum considered), because \textbf{model} is the last technique to enter overfitting for the clustering-based classifier (as shown in Figure~\ref{subfig:randcaltech101-prediction}).

The overfitting region is even more visible for \texttt{Caltech101-3} (Figure~\ref{fig:caltech101-3}).
The visual vocabulary sizes are considerably higher than for the other datasets, relative to the number of classes.
In Figure~\ref{subfig:caltech101-3-prediction} performances of all visual vocabulary techniques descend sharply for higher values of vocabulary size.
The evaluation using the SVM classifier, in Figure~\ref{subfig:caltech101-3-svm}, also clearly shows the dome-like shape.

\subsection{Influence of parameter $\alpha$}
\label{subsec:experiments-parameter-alpha}

\begin{figure}[ht]
	\centering
	
	\subfloat[]{ 
		\label{subfig:alpha-prediction}		
		\includegraphics[width=0.48\textwidth]{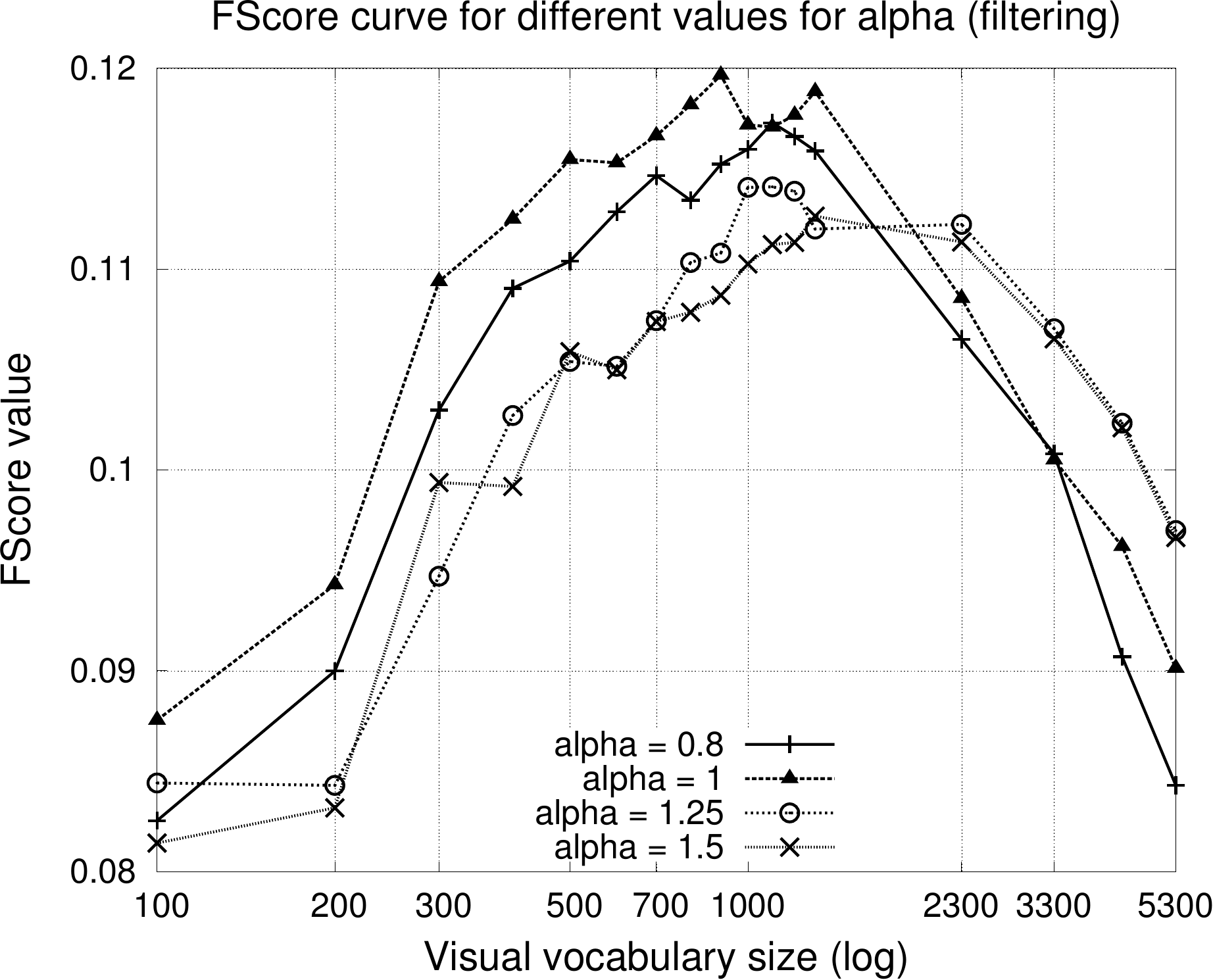} 
	}
	\hfill
	\subfloat[]{ 
		\label{subfig:alpha-svm}
		\includegraphics[width=0.48\textwidth]{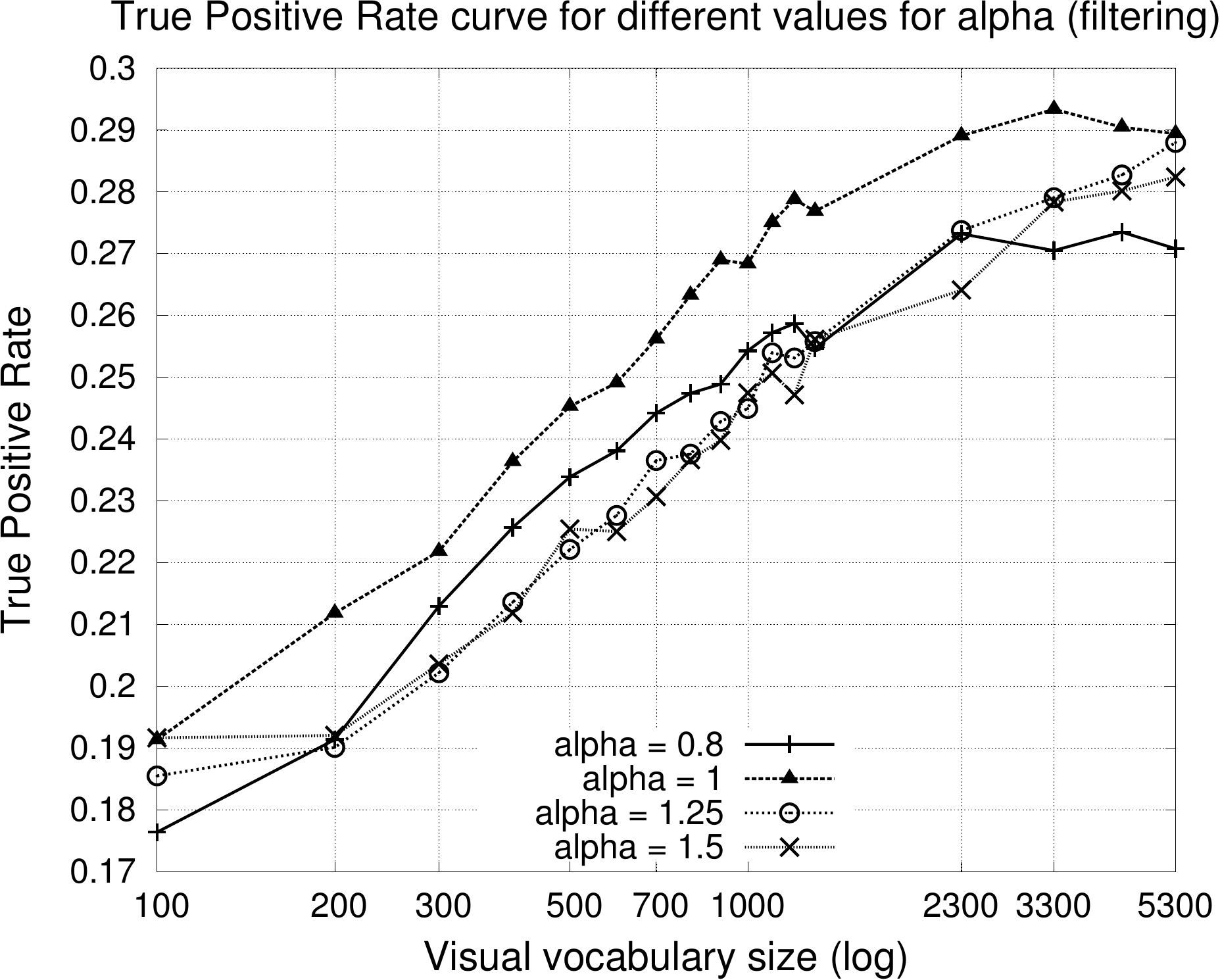} 
	}
	
	\caption{\texttt{RandCaltech101}: influence of parameter $\alpha$ on \textbf{filt+model} construction technique in the clustering-based classifier (a) and the SVM classifier (b)}
	\label{fig:parameter-alpha}
\end{figure}

In Equation~\ref{eq:threshold-alpha}, we have defined $\delta$, the filtering threshold, which is used to decide if a feature has any similar features in the known positive set.
The parameter $\alpha$ is used to fine-tune this threshold.
If $\alpha$ is set too low, only the features that are very close (in terms of Euclidean distance) are considered to be similar.
Consequently, the filtering is very strict, lowering the number of \textit{false positives}, with the risk of an inflation of \textit{false negatives}.
On the other hand, setting $\alpha$ too high allows distant features to be considered as similar, causing a high number of \textit{false positives}.
In the previous experiments, we have set the parameter $\alpha = 1$.
In this section, we study the influence of this parameter on the performances obtained by the \textbf{filt+model} construction technique.

Figure~\ref{fig:parameter-alpha} shows the evolution of the \textbf{filt+model} visual vocabulary construction technique as a function of the vocabulary size, when using $\alpha \in \{0.8,$ $1,$ $1.25,$ $1.5\}$.
The horizontal axis is logarithmic.
A value for $\alpha = 0.8$ is too strict and the high number of \textit{false negatives} decreases the classification performances.
Augmenting $\alpha = 1$ improves performances, both when using the clustering-based classifier (Figure~\ref{subfig:alpha-prediction}) and when using the SVM classifier (Figure~\ref{subfig:alpha-svm}).

If $\alpha$ is set too high, performances decrease again.
Too many features are considered similar and less features get filtered.
Performances approach those obtained when no filtering is applied.
$\alpha = 1.25$ and $\alpha = 1.5$ show similar performances, since both levels are already too high for filtering to be effective.
For $\alpha \geq 1.25$, \textbf{filt+model} is equivalent to the \textbf{model} visual vocabulary construction technique.
In Figure~\ref{subfig:alpha-prediction}, \textbf{filt+model} with $\alpha \in \{1.25,$ $1.5\}$ obtains, for high visual vocabulary sizes ($m > 2000$), better results than \textbf{filt+model} with $\alpha \in \{0.8,$ $1\}$.
This behavior is similar with that already seen in Figure~\ref{subfig:randcaltech101-prediction}, when \textbf{model} enters overfitting later than \textbf{filt+model}, and obtains better results for high vocabulary sizes.

These initial experiments make us believe that $\alpha$ is dataset independent (a value of 1 provided best results on all three datasets), but further experiments on other datasets are required for a firm conclusion.
Furthermore, a heuristic for automatically determining its value is part of our future plans.

\section{Conclusion and future work}
\label{sec:conclusions}

\paragraph{Conclusion}
In the work presented in this article, we have focused on constructing a semantically-enriched representation for images, by leveraging additional information under the form of non-positional labels.
We argue that enriching the semantics of the image representation boosts the performances of learning algorithms and we apply our proposed method to the learning task of content-based image classification.

We use the additional information in the phase of visual vocabulary construction, when building a ``bag-of-features'' image representation.
We have proposed two novel approaches for incorporating this semantic knowledge into the visual vocabulary creation.
The first approach creates dedicated vocabularies for each label, while the second uses a pre-processing phase for filtering visual features unlikely to be associated with a given object.
We have shown that the semantically-enriched image representations built using our proposals obtain higher scores than a baseline \emph{BoF} representation, in the context of a task of content-based image classification.
This shows that incorporating semantic knowledge in the vocabulary construction results in more descriptive visual words, especially on datasets where the background noise is significant.
Even when overfitted, our proposals continue to outperform the generic approach.

\paragraph{Future work}
Our visual vocabulary construction techniques, proposed in Section~\ref{sec:proposals}, are not limited to the task of object-based image classification.
They can be used with any image-related machine learning task that involves constructing a \textit{BoF} representation for images.
Whatsoever, scaling our approaches to other applications involves relaxing some of the assumptions.
For example, we assumed that labels which denote objects appear independently in the image collection.
We are working on relaxing this strict condition and on passing from a learning task of object categorization to one of scene classification.
This raises the difficulty of object co-occurrence.
For example, a picnic scene is defined by the simultaneous presence of \textit{``people'', ``trees'', ``grass''} and \textit{``food''}.
In terms of labels, this translates into label co-occurrence.
Our approaches can be scaled to image classification by addressing the label co-occurrence issue.
We are currently working on using the unsupervised feature\footnote{Note that, in this context, the word \textbf{feature} is used in the sense of attribute. 
It should not be confused with the definition of \textbf{feature} in image processing literature, where it has the sense of visual feature.} construction algorithm proposed in~\cite{RIZ13a} to reconstruct the image labels and to reduce, even eliminate, their co-occurrence.
The new labels are constructed as conjunctions of existing labels and their negations, and would actually no longer be used to label objects, but scenes.
For example, if the labels \textit{``motorcycle''} and \textit{``rider''} appear often together, a new label \textit{``motorcycle $\wedge$ rider''} will be created to mark the scene identified by the presence of the two objects.

{\small
%

}

\end{document}